\documentclass[journal]{IEEEtran}
\usepackage{cite}
\usepackage{xcolor}
\usepackage[pdftex]{graphicx}
\usepackage{amsmath}
\usepackage{amssymb}
\usepackage{bm}

\usepackage{algorithm,algpseudocode}

\usepackage{multirow}
\usepackage{caption}
\usepackage{subfigure}
\usepackage{makecell}
\usepackage{booktabs}

\usepackage{xspace}

\begin{document}
\title{InfoStyler: Disentanglement Information Bottleneck for Artistic Style Transfer}

\author{Yueming Lyu,
        Yue Jiang, 
        Bo Peng,~\IEEEmembership{Member,~IEEE},\\
        and Jing Dong,~\IEEEmembership{Senior Member,~IEEE}
\thanks{Yueming Lyu, Yue Jiang, Bo Peng, Jing Dong~(corresponding author) are with the Center for Research on Intelligent Perception and Computing~(CRIPAC), State Key Laboratory of Multimodal Artificial Intelligence Systems, Institute of Automation, Chinese Academy of Sciences~(CASIA), Beijing 100190, China, and also with the School of Artificial Intelligence, University of Chinese Academy of Sciences, Beijing 100049, China (E-mail: yueming.lv@cripac.ia.ac.cn; jiangyue2021@ia.ac.cn; bo.peng@nlpr.ia.ac.cn; jdong@nlpr.ia.ac.cn).}
}

\markboth{IEEE TRANSACTIONS ON CIRCUITS AND SYSTEMS FOR VIDEO TECHNOLOGY}%
{Shell \MakeLowercase{\textit{et al.}}: Bare Demo of IEEEtran.cls for IEEE Journals}

\maketitle

\begin{abstract}
Artistic style transfer aims to transfer the style of an artwork to a photograph while maintaining its original overall content. Many prior works focus on designing various transfer modules to transfer the style statistics to the content image. Although effective, ignoring the clear disentanglement of the content features and the style features from the first beginning, they have difficulty in balancing between content preservation and style transferring. To tackle this problem, we propose a novel information disentanglement method, named \textit{InfoStyler}, to capture the minimal sufficient information for both content and style representations from the pre-trained encoding network. InfoStyler formulates the disentanglement representation learning as an information compression problem by eliminating style statistics from the content image and removing the content structure from the style image. Besides, to further facilitate disentanglement learning, a cross-domain Information Bottleneck (IB) learning strategy is proposed by reconstructing the content and style domains. Extensive experiments demonstrate that our InfoStyler can synthesize high-quality stylized images while balancing content structure preservation and style pattern richness. 
\end{abstract}


\label{sec:intro}
\IEEEPARstart{A}{rtistic} style transfer aims at transferring the style of a reference image onto a given content image. 
It has attracted widespread attention from academic, industrial and art communities given its significance for real applications. In recent years, researchers have applied convolutional neural networks (CNNs) for image translation and stylization~\cite{zhang2019image,tan2020incremental,gao2021wallpaper,fu2021let}. Starting from \cite{gatys2016image,johnson2016perceptual}, existing approaches can be divided into two typical categories: optimization-based and feedforward network-based approaches.

Optimization-based approaches~\cite{gatys2016image,kolkin2019style,risser2017stable} leverage iterative optimization for per image style transfer, which are complex and have high computational cost. 
In contrast, feedforward network-based methods\cite{chen2016fast,johnson2016perceptual,chen2017stylebank,li2017universal,li2019learning,sheng2018avatar,park2019arbitrary} are time-efficient. 
They mainly apply \textbf{encoder-transfer-decoder structures} as the feedforward network and infer stylized images using the network directly. 
The structures of such design usually consist of \textbf{three stages}: 
\textbf{1)} the first stage is a fixed encoder (\emph{e.g.}, pre-trained VGG-19~\cite{simonyan2014very}) for extracting content and style features.
\textbf{2)} the second stage is a well-designed transfer module to transfer the style statistics from the style features to the content features.
\textbf{3)} A learned decoder, as the third stage, outputs realistic stylized images.
\begin{figure*}[t!]
	\centering
	  \includegraphics[width=0.8\linewidth]{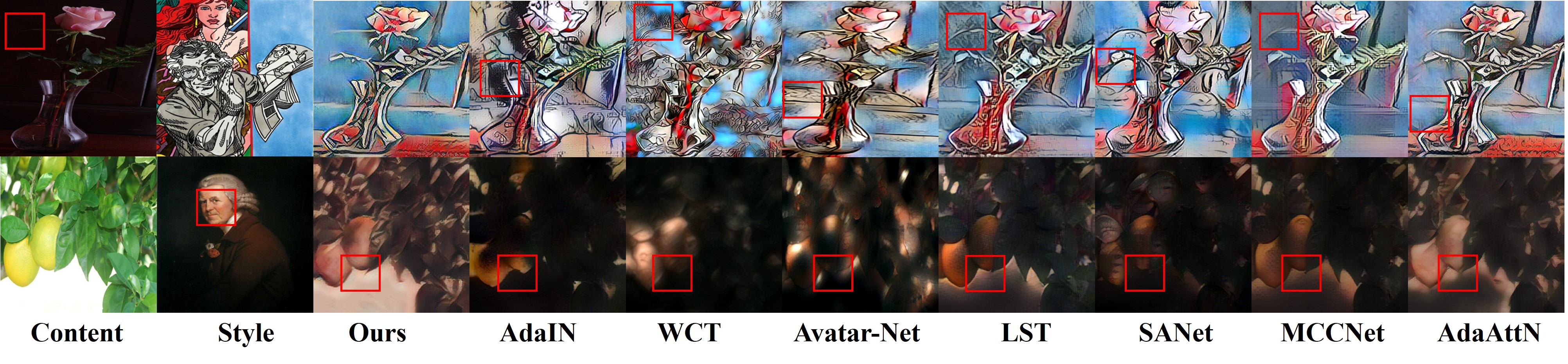}
	  \caption{Comparisons with state-of-the-art style transfer methods. Our method is outstanding in terms of content preservation and texture stylization. Specifically, our method preserves the clear content structure without leaking the color distribution of the content image. This is in contrast to other methods that leak the black background color of the content image into the stylized results (as shown in the first row with red boxes). On the other hand, when rich styles are mostly concentrated in a local region of the style image, our method can accurately transfer style information from the given style. However, previous methods lead to undesired and messy outputs, resulting in damaged content structure (as shown in the second row with red boxes).}
    \label{fig:shouye}
\end{figure*}

Existing feed-forward algorithms~\cite{chen2017stylebank,huang2017adain,li2017universal,li2019learning,sheng2018avatar,park2019arbitrary,liu2021adaattn,wu2021styleformer} focus on designing more efficient and effective transfer modules in the second stage, but they overlook the importance of representation optimization in the first stage, leading to unsatisfactory results, as shown in Figure~\ref{fig:shouye}. 
\textbf{The content information, such as edges and contours, and the style information, such as textural patterns and color patterns}, are entangled in both the content image and the style image.
Extracting them using the same pre-trained encoding network indiscriminately would make it difficult to achieve robust and accurate stylized results.
As a result, the following \textbf{questions} are not yet explicitly answered by current methods: 
\textbf{1)} \textit{what to preserve in the content image?}
A good stylized effect should only preserve the original content structure without leaking the styles of the content image.
For example, in the first row of Figure~\ref{fig:shouye}, the background color of the content image leaks into the stylized results (4th, 8th, and 10th columns), which is not a desirable phenomenon. 
\textbf{2)} \textit{where to look in the style image?}
When styles reside in a relatively small region of the style image (\emph{e.g.} the head region of the second-row style image), the final output should draw more style information from that informative region instead of from other large but uninformative regions. 
Without this consideration, as shown in the second row, existing methods tend to generate disharmonious styles biased by the large black background.

\begin{figure}[t!]
	\centering
	  \includegraphics[width=0.8\columnwidth]{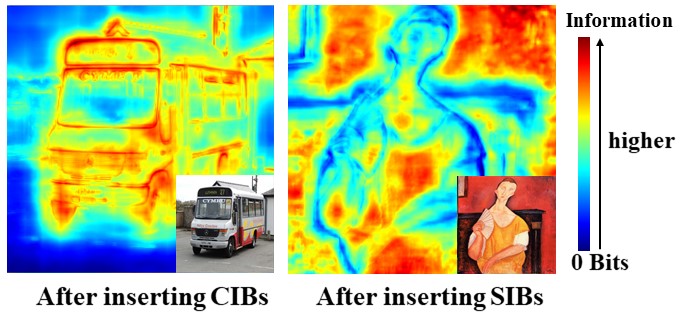}
	  \caption{Visualization of the contained information ~(quantified in bits) in the internal content features and style features after inserting the proposed Content Information Bottlenecks (CIBs) and Style Information Bottlenecks (SIBs) respectively. 
    It indicates that the content features are compressed to only structural information~(\emph{e.g.} edges and contours) while the style features are optimized to better focus on style information~(\emph{e.g.} colors and textures).}
    \label{fig:vis}
  \end{figure}

To tackle the problems mentioned above, we propose a novel approach, called \textbf{\emph{InfoStyler}}, to optimize the representation from the first beginning of the feedforward network and capture disentangled content and style representations from the perspective of information theory~\cite{cernekova2005information,shwartz2017opening,achille2018emergence}. 
We formulate the disentanglement representation learning as an information compression problem and disentangle content-related features and style-related features explicitly. 
Specifically, we introduce Content Information Bottlenecks (CIBs) and Style Information Bottlenecks (SIBs) to the pre-trained encoding network in the first stage of the current feedforward network-based pipeline.
By explicitly supervising the flow of information, the CIBs are designated to eliminate the style statistics from the content image and preserve only structural information. On the contrary, the SIBs learn to remove the content structure from the given style image and only capture its style information. Figure~\ref{fig:vis} shows the visualization of the contained information after CIBs and SIBs optimization.

To better facilitate disentanglement learning, we additionally design a cross-domain Information Bottleneck (IB) learning strategy for our style transfer model. 
By cross-exchanging the input domains of the proposed CIBs and SIBs, and then reconstructing the style image and the content image in a symmetric cycle manner, we can introduce more supervision for the style transfer learning.
In this way, the transferred results are able to reflect more accurate and disentangled content information and style information from the respective input images. 

We summarize our contributions as follows:
\begin{itemize}
\item We introduce information bottleneck to improve the feature representation capability for artistic style transfer.
The proposed Content Information Bottlenecks (CIBs) and Style Information Bottlenecks (SIBs) can disentangle content-related features and style-related features explicitly. 
To the best of our knowledge, we are the first to successfully introduce this information-theoretic tool to style transfer.
\item To further improve the representation disentanglement, we introduce a cross-domain IB learning strategy. 
By cross-exchanging the input domains of CIBs and SIBs, it prompts the network to extract more accurate and disentangled content and style information. 
\item Extensive experiments demonstrate that the proposed InfoStyler can disentangle information, balance content preservation and texture stylization, and generate high-quality stylized images.
\end{itemize}

\section{related work}
\subsection{Artistic Style Transfer}

Artistic style transfer aims at transferring artistic styles from style images to given content images.
It has been studied for several years. 
Traditional approaches utilized several techniques based on image filtering~\cite{winnemoller2006real}, image analogy~\cite{hertzmann2001image, frigo2016split, song2017learning, song2019joint}, and stroke rendering~\cite{hertzmann1998painterly} to perform style transfer. 
With advances in deep neural networks in recent years, numerous neural style transfer methods have been introduced.
These methods can be divided into two categories: optimization-based methods and feedforward network-based methods.
For optimization-based methods, Gatys \emph{et al.}~\cite{gatys2015texture} first proposed a Gram loss upon deep features from a pre-trained neural network (\emph{e.g.} VGG~\cite{simonyan2014very}) to represent the styles of images. 
After that, a number of variants~\cite{ruder2016artistic,champandard2016semantic,risser2017stable,li2017demystifying,gatys2017controlling,kolkin2019style,du2020much} were developed to improve visual quality based on new style and content losses, or extended to different scenarios and domains. 
Specifically, Kolkin \emph{et al.}~\cite{kolkin2019style} proposed a new style loss to measure the distance between style features and content features by Earth Movers Distance~(rEMD). 
Gatys \emph{et al.} further extended their prior work~\cite{gatys2015texture} to controlled stylization~\cite{gatys2017controlling}, which included control over spatial location, color information and across spatial scale.  
However, the optimization-based methods have high computational complexity, and thus are not suitable for real-time applications. 

Different from optimization-based methods, feedforward network-based methods are time-efficient, and can be trained for real-time style transfer. 
Li \emph{et al.}~\cite{li2017universal} introduced the whitening and coloring transform (WCT) to directly match content feature statistics with given style images. 
Huang \emph{et al.}~\cite{huang2017adain} proposed adaptive instance normalization (AdaIN) to align the mean and variance of the content features with the style features.
Li \emph{et al.}~\cite{li2019learning} formulated the image style transfer problem as a linear transformation between the content and style features. 
Chen \emph{et al.}~\cite{chen2016fast} proposed an optimization objective based on local matching that combined content structures and style textures in feature space.
Sheng \emph{et al.}~\cite{sheng2018avatar} proposed AvatarNet to transfer the content features to semantically nearest style features by a patch-based style decorator.  
Park \emph{et al.}~\cite{park2019arbitrary} introduced SANet to efficiently integrate local style patterns according to the semantic spatial distribution of the content image. 
Deng \emph{et al.}~\cite{deng2021arbitrary} proposed Multi-Channel Correlation network (MCCNet), which aligned features to input via per-channel correlation to render relative clean stylized outputs.
Liu \emph{et al.}~\cite{liu2021adaattn} designed an Adaptive Attention Normalization (AdaAttN) and proposed a local feature loss to enhance local visual quality. 
Lin \emph{et al.}~\cite{lin2021drafting} proposed a two-stage architecture including a Drafting Network to capture global style features in low-resolution and a Revision Network to revise the local details in high-resolution. 
Wu \emph{et al.}~\cite{wu2021styleformer} introduced a transformer-inspired feature-level method and adopted the multi-head attention to globally model new style codes for the generated images. 
Deng \emph{et al.}~\cite{deng2022stytr2} proposed a transformer-based approach to enhance long-range dependencies and capture global information of input images for image style transfer.

\textbf{\emph{Discussions about some works on image-to-image translation.}}
Our proposed framework indeed shares some similarities with MUNIT~\cite{huang2018multimodal}, DRIT~\cite{lee2018diverse}/DRIT++~\cite{lee2020drit++}, as both methods aim to perform unpaired image translation and are based on cycle consistency to provide more supervision for the transfer process.
However, there are some key differences between our method and other works such as MUNIT, and DRIT/DRIT++.  
Firstly, MUNIT and DRIT/DRIT++ are based on the assumption that an image can be represented into a domain-invariant content space that captures shared content information across domains and a domain-specific style space for each domain, which means they can only perform image translation between two visually similar domains (\emph{e.g.}, horses$\leftrightarrow$zebras). Differently, Our method is based on the assumption that an image can be decomposed into content-related information (\emph{e.g.}, edges and contours) and style-related information (\emph{e.g.}, textural patterns and color patterns). Thus, InfoStyler does not require the content image and style image to be similar in content. 
Secondly, MUNIT and DRIT/DRIT++ enforce cycle consistency by cross-exchanging the content codes extracted from different domains. Differently, the proposed cross-domain IB learning involves cross-exchanging the input domain images sent to the encoders, namely extracting the content information from the style domain and the style information from the content domain.

\textbf{\emph{Learning disentangled representations.}}
Disentanglement is one of the key ideas that make style transfer more natural and accurate since in the task of style transfer, style and content features should be disentangled due to the domain deviation. 
Specifically, Kotovenko et al.~\cite{kotovenko2019content} proposed a fixpoint triplet loss and a disentangle loss to minimize the distance between different stylized results with different content images and a common style image.
Deng et al.~\cite{deng2020arbitrary} introduced a content self-adaptation module and a style self-adaptation module to adaptively disentangle the content and style representations. Additionally, they proposed a disentangle loss that also constrains the content features extracted from a series of stylized results generated using the same content image but different style images to be similar, as well as the style features extracted from a series of stylized results generated using the same style image but different content images to be similar. However, due to lack of explicit supervision, it is still challenging to learn well-disentangled representations. In this paper, we propose an information disentanglement method to disentangle the extracted content and style features from the aspect of information compression.

\begin{figure*}[t]
	\begin{center}
		\includegraphics[width=0.8\linewidth]{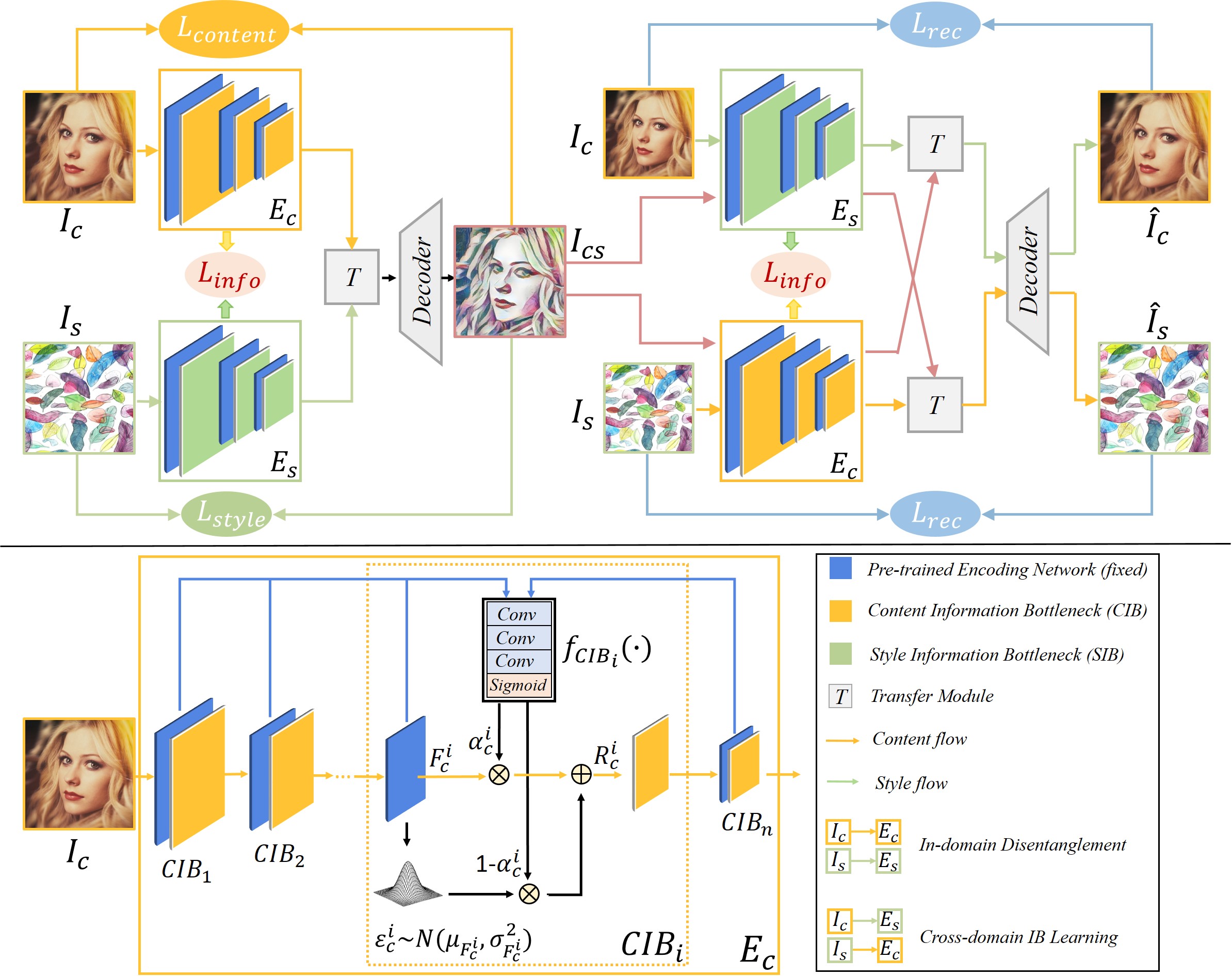}
	\end{center}
	\caption{An overview of the architecture of our method. The top figure is the overall pipeline of our InfoStyler. First, take the content image $I_c$ and the style image $I_s$ as input, the encoding networks $E_c$ and $E_s$ equipped with CIBs and SIBs extract disentangled content features and style features respectively. Then, the transfer module and the decoder output the stylized result $I_{cs}$. After that, the inputs of $E_c$ and $E_s$ are cross-exchanged to form a symmetric cycle reconstruction process where reconstructed content and style images, $\hat{I}_c$ and $\hat{I}_s$, are produced. 
  The bottom figure shows the details of $E_c$, which extracts content information and eliminates style statistics by adding noise. The details of $E_s$ are similar to $E_c$, but $E_s$ is used to remove the content information and capture the style information. Best viewed in color.
  }
\label{fig:method}
\end{figure*}
\subsection{Information Bottleneck} 
The seminal work of Tishby \emph{et al.}~\cite{tishby2015deep} first analyzed DNNs via the theoretical framework of the information bottleneck (IB) principle. 
They suggested that the goal of the deep neural network war to optimize the information bottleneck tradeoff between compression and prediction for each layer. 
Following up on the idea of \cite{tishby2015deep}, Shwartz \emph{et al.}~\cite{shwartz2017opening} demonstrated the effectiveness of the information plane visualization of DNNs to better understand the training dynamics, learning processes, and internal representations. 
After that, many methods introduced the information bottleneck theory into their schemes to learn more effective representations.

Schulz \emph{et al.}~\cite{schulz2019restricting} adopted the information bottleneck to restrict the flow of the information and removed the unimportant information for network layers. 
DICE~\cite{rame2020dice} proposed a classification loss with information bottleneck and adversarially minimized the conditional redundancy between features. 
Since weakly supervised semantic segmentation (WSSS) needs pixel-level localization from class labels, IWSSS~\cite{lee2021reducing} reduced the information bottleneck and improved the quality of localization maps. 
IB-GAN~\cite{jeon2021ib} proposed a GAN framework with information bottlenecks for unsupervised disentangled representation learning. 
InfoSwap~\cite{gao2021information} was proposed to use the information bottleneck theory to disentangle the identity and the perceptual information for face swapping. 
Sun \emph{et al.}~\cite{sun2022information} proposed a mutual-information regulariser to disentangle identity and expression from 3D input faces. 
In this paper, we explicitly disentangle the content information and the style information in style transfer based on the information bottleneck theory. 

\section{Proposed Method}
In this section, we start with a detailed introduction of the information bottleneck theory as the preliminary in Sec.~\ref{sec:ib}. 
We then give a detailed description of the proposed method in Sec.~\ref{sec:indomain} and Sec.~\ref{sec:crossdomain}. 
Finally, we introduce the loss functions in Sec.~\ref{sec:fuc}.
\subsection{Preliminary on Information Bottleneck}
\label{sec:ib}
Information Bottleneck~\cite{tishby2000information} is an appealing information theoretic method, which describes a limitation on information redundancy. 
The goal of information bottleneck~\cite{tishby2000information, alemi2016deep,peng2018variational} is to learn a compressive representation $R$ that is maximally informative about the output $Y$, while containing as less information as possible about the input $X$.
Namely, $R$ is a \emph{minimal sufficient statistics}~\cite{fisher1922mathematical} of $X$ for producing $Y$. 
Therefore, finding an optimal representation $R$ can be formulated as the minimization of the following objective function:
\begin{equation}
    \mathcal{L}_{IB}:= I(X, R)-\beta I(Y, R) \text{,}
\label{eq:ib}
\end{equation}
where $I(\cdot , \cdot)$ denotes the mutual information, and $\beta \in [0,1]$ is a trade-off parameter between encouraging $R$ to be predictive of $Y$ and encouraging $R$ to ``forget'' $X$.
To this end, the compressed representation preserves the most relevant information about the output and eliminates the irrelevant information which does not contribute to the output.

In style transfer, the first step is to extract content information from the content image and extract style information from the style image. 
However, extracting both features from the same encoding network indiscriminately would inevitably lead to the entanglement of the content information and the style information. 
Therefore, how to extract accurate and disentangled feature representations plays a vital role in the transfer process. 
Here, we introduce multiple information-bottleneck layers to the pre-trained encoding network to eliminate the style information contained in the content features and reduce the content information in the style features. 
Our transfer model with information bottlenecks is presented in the following. 

\subsection{In-domain Disentanglement via IB}
\label{sec:indomain}
Let $\mathcal{C}$ and $\mathcal{S}$ be the domains of content images and style images, respectively. 
Our goal is to learn disentangled content and style representations from $I_c \in C$ and $I_s \in S$, and then transfer the styles to the content image to create the stylized image $I_{cs}$. 
The overall pipeline of our method is shown in Figure~\ref{fig:method}.

We employ the pre-trained VGG-19~\cite{simonyan2014very} network as the basic encoding network, $E_{0}$, which is fixed during training. 
Then, by equipping it with Content Information Bottlenecks~(CIBs) and Style Information Bottlenecks~(SIBs), we derive the effective content encoding network $E_c$ and style encoding network $E_s$, respectively. 
In this way, we can take full advantage of the powerful representation ability of the pre-trained network and eliminate the irrelevant information by the proposed information bottlenecks simultaneously. 

Specifically, taking $I_c$ and $I_s$ as input, we extract multiple levels of original features $F_c = \{F_c^1, F_c^2, \cdots, F_c^n\}$ and $F_s = \{F_s^1, F_s^2, \cdots, F_s^n\}$ from $E_{0}$
, where $n$ denotes the total number of extracted features from shallow to deep layers.
Since features $F_c$ and $F_s$ contain redundant information, we employ a multi-level strategy by integrating $n$ Content Information Bottlenecks $CIBs = \{CIB_1, CIB_2, \cdots, CIB_n\}$ and $n$ Style Information Bottlenecks $SIBs = \{SIB_1, SIB_2, \cdots, SIB_n\}$ in multiple layers.
Here, $CIBs$ focus on learning the most compact representations of the content information from $I_c$, while $SIBs$ aim to learn the content-irrelevant style information from $I_s$. 
The process of reducing information is achieved by predicting multi-level informative controllers $\alpha_c=\{\alpha_c^1, \alpha_c^2, \cdots, \alpha_c^n\}$ and $\alpha_s=\{\alpha_s^1, \alpha_s^2, \cdots, \alpha_s^n\}$ for controlling the remained information in content features and style features. 

Take the $i$-th layer as an example, $CIB_i$ and $SIB_i$ control the information available passed to the next layer.
Based on the informative controllers $\alpha_c^i$ and $\alpha_s^i$ that indicate the feature importance for producing $I_{cs}$, the information is restrained by adding noises~\cite{kingma2013auto,alemi2016deep}.
We replace the unimportant information contained in the original features with Gaussian noises, $\varepsilon_c^i$ and $\varepsilon_s^i$, by
\begin{equation}
  \begin{aligned}
    R_c^i &= \alpha_c^i F_c^i + (1 - \alpha_c^i)\varepsilon_c^i,\\
    R_s^i &= \alpha_s^i F_s^i + (1 - \alpha_s^i)\varepsilon_s^i,
  \end{aligned}
\end{equation}
where the noises~(\emph{i.e.}, $\varepsilon_c^i$ and $\varepsilon_s^i$) and the informative controllers~(\emph{i.e.,} $\alpha_c^i$ and $\alpha_s^i$) all have the same spatial dimensions as $F_c^i$ and $F_s^i$. 
In order to preserve the distribution of the $i$-th features, the noises are sampled from Gaussian distribution with the same mean and variance of the original $i$-th features, \emph{i.e.}
$\varepsilon_c^{i} \sim \mathcal{N}\left(\mu_{F_c^{i}}, \sigma_{F_c^{i}}^{2}\right)$ and $\varepsilon_s^{i} \sim \mathcal{N}\left(\mu_{F_s^{i}}, \sigma_{F_s^{i}}^{2}\right)$.
In this way, the new compressed features have the same overall distribution as the original features.

As the key component of $CIB_i$, the $i$-th content controller $\alpha_c^i$ is predicted by $f_{CIB_i}(\cdot)$ composed of convolution and sigmoid operation with all $n$ intermediate features $F_c$ as input.
The $i$-th style controller $\alpha_s^i$, as the key component of $SIB_i$, is predicted by $f_{SIB_i}(\cdot)$ in a similar way. 
They are obtained by
\begin{equation}
  \begin{aligned}
    \alpha_c^i &= f_{CIB_i}(F_c),\\
    \alpha_s^i &= f_{SIB_i}(F_s),
  \end{aligned}
\end{equation}
where all input features in $F_c$ and $F_s$ are first resized to the same size of the $i$-th features and then concatenated along the channel dimensions.  
The output $\alpha_c^i$ and $\alpha_s^i$ are of the same size as $F_c^i$ and $F_s^i$, with values in $[0,1]$ after \emph{sigmoid} operation.

After that, the multiple compressed features $R_c= \{R_c^1, R_c^2, \cdots, R_c^n\}$ and $R_s= \{R_s^1, R_s^2, \cdots, R_s^n\}$ are obtained.
For supervising the information disentanglement and compression process, the mutual information before and after injecting noises are evaluated to optimize the compressed features, which will be elaborated in Sec.~\ref{sec:fuc} in the Information Loss part. 

Finally, we send the disentangled and compressed features into the transfer module and the decoder to output the stylized images.
Following \cite{liu2021adaattn}, we adopt the Adaptive Attention Normalization~(AdaAttN) module as the transfer module $\mathcal{T}(\cdot, \cdot)$.
AdaAttN is a multi-level attention-based transfer module that enables a point-by-point transfer of feature distributions by utilizing both shallow and deep features with an attention mechanism, where more details of the AdaAttN module can be seen in \cite{liu2021adaattn}. 
A symmetric structure of VGG-19 is used as the decoder $Dec(\cdot)$.
The stylized image $I_{cs}$ is synthesized by
\begin{equation}
    I_{cs} = Dec(\mathcal{T}(R_c, R_s)).
\end{equation}

\subsection{Cross-domain IB Learning}
\label{sec:crossdomain}
To better facilitate disentanglement learning, we introduce the cross-domain IB learning strategy in a self-supervised manner. 
Inspired by CycleGAN~\cite{zhu2017unpaired}, cross-domain IB learning is formulated as a symmetric cycle reconstruction that can provide pixel-wise supervision for the transfer process and thus can produce more high-quality stylized results with richer styles and clearer content structure. 
Specifically, $E_c$ extracts content information from $I_s \in \mathcal{S}$ and $E_s$ extracts style information from $I_c \in \mathcal{C}$, which is contrary to the input domains of in-domain disentanglement learning. 
Then, the cross-domain features $R_c^{(s)}$ and $R_s^{(c)}$ can be obtained by
\begin{equation}
  \begin{aligned}
  R_c^{(s)} = E_c(I_s) ,\\
  R_s^{(c)} = E_s(I_c) .
  \end{aligned}
\end{equation}
By cross-exchanging the input domains of $E_c$ and $E_s$, the CIBs are assigned to extract content information from the original style image domain $\mathcal{S}$, while the SIBs are used to extract style information from the original content image domain $\mathcal{C}$.
Namely, the content information, such as content structures, is extracted from the artworks images, while the style information, \emph{e.g.} colors and textures, is obtained from photo images.
After that, the additional content and style features from the generated image $I_{cs}$ are extracted by
\begin{equation}
  \begin{aligned}
    R_c^{(cs)} = E_c(I_{cs}),\\
    R_s^{(cs)} = E_s(I_{cs}). 
  \end{aligned}
\end{equation}

Therefore, with $R_c^{(cs)}$ and $R_s^{(cs)}$ that contain the content and style information of $I_{cs}$,
$R_c^{(s)}$ and $R_s^{(c)}$ that contain the content and style information of $I_s$ and $I_c$,
the reconstructed results $\hat{I}_c$ and $\hat{I}_s$ can be inferred by sending the features into the transfer module and the decoder via the following process:
\begin{equation}
    \begin{aligned}
    \hat{I}_c = Dec(\mathcal{T}(R_c^{(cs)}, R_s^{(c)})) , \\
    \hat{I}_s = Dec(\mathcal{T}(R_c^{(s)}, R_s^{(cs)})) .
    \end{aligned}
\end{equation}
\begin{figure*}[t]
	\begin{center}
		\includegraphics[width=0.75\linewidth]{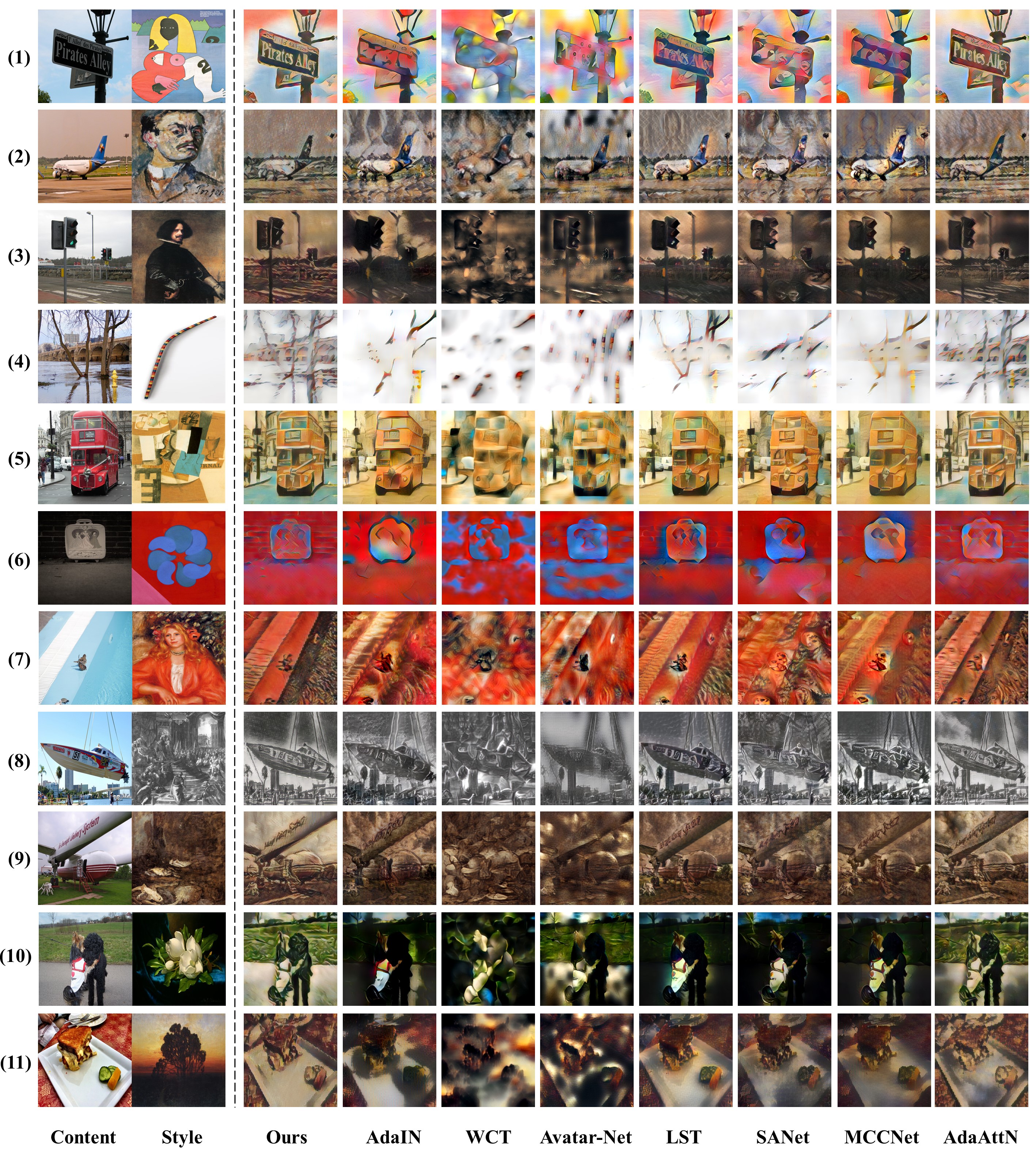}
	\end{center}
	\caption{Qualitative comparisons with seven state-of-the-art methods.}
	\label{fig:qualitative}
\end{figure*}
In this way, $R_c^{(cs)}$ has as much content information from the original $I_c$ as possible, and $R_s^{(cs)}$ has as much style information from the original $I_s$ as possible. 
Therefore, the encoding network with CIBs and SIBs has more supervision to learn distinctive content and style representation, and a better stylized image $I_{cs}$ can be produced.
Note that the trainable networks in cross-domain IB learning share weights with in-domain disentanglement learning.

\subsection{Loss Functions}
\label{sec:fuc}
\textbf{Information loss.}
To add supervision during the information compression process, we utilize the mutual information $MI(R^i, F^i)$ to compute the information shared between $F^i$ and $R^i$, where $F^i$ and $R^i$ can either be from the content image or the style image. 
The information loss is formulated as 
\begin{equation}
\label{eq:info}
    L_{info}=\frac{1}{n} \sum_{i=1}^{n} MI\left(R^i, F^i\right).
\end{equation}
Since there is no ground-truth for style transfer, we only use the first term in Eq.~(\ref{eq:ib}) as the explicit supervision for the proposed information bottlenecks which is fullfilled by Eq.~(\ref{eq:info}) here. The second term in Eq.~(\ref{eq:ib}) can be implicitly represented by the following content loss (Eq.~(\ref{con:content})) and Style loss (Eq.~(\ref{con:style})). This is because the content loss and style loss both strive to maximize the mutual information between the output Y and the intermediate original feature F, while the designed information loss aims to minimize the mutual information between the optimized feature R and the intermediate original feature F. Therefore, the joint effect of these losses will implicitly increase the mutual information between the target Y and the optimized feature R.
To simplify the calculation of $MI$~\cite{doersch2016tutorial,schulz2019restricting, gao2021information}, we normalize the distribution of $R^i \mid F^i$ and $R^i$ by their mean and variance since the KL-divergence is invariant under scaling.

\begin{table*}[!t]
  \centering
  \caption{The quantitative comparison and execution time analysis with different methods. The results of user preference include three aspects, \emph{i.e.}, content preservation, texture stylization, and overall performance.}
  \label{tab:quantitative}
  \begin{tabular}{cc|c|c|c|c|c|c|c|c}
  \toprule
  \multicolumn{2}{c|}{Method}& AdaIN  & \multicolumn{1}{l|}{WCT} & \multicolumn{1}{l|}{Avatar-Net} & \multicolumn{1}{l|}{LST} & \multicolumn{1}{l|}{SANet} & \multicolumn{1}{l|}{MCCNet} & \multicolumn{1}{l|}{AdaAttN} & \multicolumn{1}{l}{Ours} \\ 
  \hline
  \hline
  \multicolumn{1}{c|}{\multirow{3}{*}{Preference (\%)}}& Content Preservation &2.0 &1.3 &2.3 &22.8 &2.3&8.4 &21.6 & \textbf{39.3} \\ \cline{2-10} 
  \multicolumn{1}{c|}{}& Texture Stylization  & 6.3 & 10.6 & 9.3 & 15.0 & 9.1 & 10.7 & 13.6 &\textbf{25.4}\\ \cline{2-10} 
  \multicolumn{1}{c|}{}& Overall Performance  & 2.7 & 2.3  & 3.6  & 20.1 & 3.7 & 8.7  & 21.7 & \textbf{37.2}\\ 
  \hline
  \multicolumn{2}{c|}{Content loss $\downarrow$}&4.739  &6.486 & 5.845 & 4.087 & 4.536& 4.339 & 4.214 & \textbf{4.028}\\ \hline
  \multicolumn{2}{c|}{SSIM $\uparrow$} & 0.430  & 0.368 & 0.476& 0.524  & 0.445 & 0.524 & 0.506&\textbf{0.533} \\ \hline
  \multicolumn{2}{c|}{Gram loss $\downarrow$} & 0.0174 & 0.0125  & 0.0447& 0.0151 & 0.0147 & 0.0154 & 0.0159& 0.0146 \\ 
  \hline
  \multicolumn{1}{c|}{\multirow{2}{*}{\begin{tabular}[c]{@{}c@{}}Inference time\\ (sec./image)\end{tabular}}} & 256$\times$ 256 & 0.005  & 0.601  & 1.223   & 0.017   & 0.009  & 0.013 & 0.051 & 0.121 \\ \cline{2-10} 
  \multicolumn{1}{c|}{} & 512$\times$ 512 & 0.006  & 1.283  & 2.874& 0.038 & 0.014 & 0.020 & 0.112 & 0.430 \\ 
  \bottomrule
  \end{tabular}
  \end{table*}

Thus the mutual information can be simplified as
\begin{equation}
  \begin{aligned}
      MI\left(R^i, F^i\right) \triangleq&D_{KL}\left[P\left(R^i \mid F^i\right) \| P\left(R^i\right)\right] \\
      \approx&-\frac{1}{2}\left[1-\left(\alpha^i \frac{F^i-\mu_{F^i}}{\sigma_{F^i}}\right)^{2}-\left(1-\alpha^i\right)^{2}\right] \\
      &-\log(1-\alpha^i),
  \end{aligned}
\end{equation}
where $D_{KL}[\cdot\|\cdot]$ represents the KL-divergence between two distributions, $P\left(R^i \mid F^i\right)$ and $P\left(R^i\right)$ denote the respective probability distributions. The deviation process is shown in Appendix.

\textbf{Content loss.}
We adopt the local feature loss proposed by \cite{liu2021adaattn} as our content loss, which constrains the result content structure to be similar to the input content image.
It constrains the features of generated image to be consistent with the transferred result by the transfer module \emph{AdaAttN}~\cite{liu2021adaattn}:
\begin{equation}
    \mathcal{L}_{content}=\sum_{i=1}^{n}\left\|E_{0}^i\left(I_{cs}\right)-\mathcal{T}_i^{\prime}\left(F_{c}, F_{s}\right)\right\|_{2},
\label{con:content}
\end{equation}
where $E_{0}^i$ denotes the pre-trained VGG encoder to extract $i$-th layer features. 
Here, $\mathcal{T}^{\prime}_i$ is the parameter-free version of \emph{AdaAttN} without learnable parameters, which provides an initial transfer result as good supervision in the content aspect.

\textbf{Style loss.}
For style loss, we adopt the commonly used mean-variance loss~\cite{huang2017adain}, which evaluates the distances of mean $\mu$ and variance $\sigma$ between the generated image and the style image in VGG feature space to obtain a stylized effect. 
It can be formulated as
\begin{equation}
    \begin{aligned}
    \mathcal{L}_{style}=& \sum_{i=1}^{n}\left(\left\|\mu\left(E_{0}^i\left(I_{cs}\right)\right)-\mu\left(E_0^i(I_s)\right)\right\|_{2}\right.\\
    &\left.+\left\|\sigma\left(E_{0}^i\left(I_{cs}\right)\right)-\sigma\left(E_0^i(I_s)\right)\right\|_{2}\right).
    \end{aligned}
\label{con:style}
\end{equation}

\textbf{Reconstruction loss.}
For the reconstruction process in cross-domain IB learning strategy, we adopt content image and style image reconstruction loss, which is defined as
\begin{equation}
    \begin{aligned}
    \mathcal{L}_{rec} = \|\hat{I}_c - I_c\|_{2} + \|\hat{I}_s - I_s\|_{2}. \\
    \end{aligned}
\end{equation}

\textbf{Total loss.} In summary, the full objective function is:
\begin{equation}
    \mathcal{L} = \lambda_{i}\mathcal{L}_{info} + \lambda_{c}\mathcal{L}_{content} + \lambda_{s}\mathcal{L}_{style} + \lambda_{r}\mathcal{L}_{rec} \text{,}
\label{con:totalloss}
\end{equation}
where $\lambda_i$, $\lambda_l$, $\lambda_s$, $\lambda_r$ are trade-off parameters of different loss terms. 

\section{Experiments}
\subsection{Implementation Details}

To train the model, we use the content images from \emph{MS COCO}~\cite{lin2014microsoft} dataset and the style images from \emph{WikiArt}~\cite{karayev2013recognizing} dataset, respectively.
For the optimizer, we choose ADAM~\cite{kingma2015adam} with $\beta_{1}$ = 0.9, $\beta_{2}$ = 0.999.
During training, we first resize all images to 512 $\times$ 512 resolution and then randomly crop them to 256 $\times$ 256 patches for augmentation. 
We use a batch size of 2 and train the model on 1 NVIDIA TITAN RTX GPU. 
In experiments, the trade-off parameters in Eq.~(\ref{con:totalloss}) are set as $\lambda_{i}=5, \lambda_{c}=3, \lambda_{s}=10, \lambda_{r}=10$.
The total number $n$ of optimized features in this paper is set to 4, including $Relu\_2\_1, Relu\_3\_1, Relu\_4\_1, Relu\_5\_1$ of the VGG-19. 

\subsection{Comparison with Previous Methods}

\subsubsection{\textbf{Qualitative Comparison}}
As shown in Figure~\ref{fig:qualitative}, we compare our method with seven previous state-of-the art style transfer methods, \emph{i.e.}, AdaIN~\cite{huang2017adain}, WCT~\cite{li2017universal}, Avatar-Net~\cite{sheng2018avatar}, LST~\cite{li2019learning}, SANet~\cite{park2019arbitrary}, MCCNet~\cite{deng2021arbitrary}, and AdaAttN~\cite{liu2021adaattn}.
AdaIN~\cite{huang2017adain} simply adjusts the mean and variance of content features, thus providing a sub-optimal solution with many content details lost (1st, 3rd and 11th rows). Although WCT~\cite{li2017universal} optimally matches the second-order statistics, it often yields messy stylized results (2nd, 3rd, and 7th rows). Since Avatar-Net~\cite{sheng2018avatar} directly swaps the style patches to the corresponding content patches, it may lead to blurry stylized results with artifacts (1st, 2nd, and 10th rows). LST~\cite{li2019learning} focuses on transferring the lower-level style patterns and usually produces less stylized images (1st, 2nd, and 4th rows), and even remains content image color (2nd row). As an attention-based method, SANet~\cite{park2019arbitrary} is more likely to apply repeating style patterns into the content images improperly (2nd, 3rd, and 11th rows). 
In the 7th row, SANet~\cite{park2019arbitrary} even transfers some style patches into the results directly. 
MCCNet~\cite{deng2021arbitrary} and AdaAttN~\cite{liu2021adaattn} both produce relatively clean stylized results. However, MCCNet blurs the content structures in many cases (1st, 7th, 8th row) and AdaAttN~\cite{liu2021adaattn} produces uneven style patterns in the stylization results (3rd, 5th, 10th, and 11th row).
Compared with these approaches, our method achieves a better balance between style pattern richness and content structure consistency, thanks to the novel disentangled information bottlenecks, which can be observed in Figure~\ref{fig:qualitative}. 
 
\begin{figure}[t]
	\begin{center}
		\includegraphics[width=0.8\columnwidth]{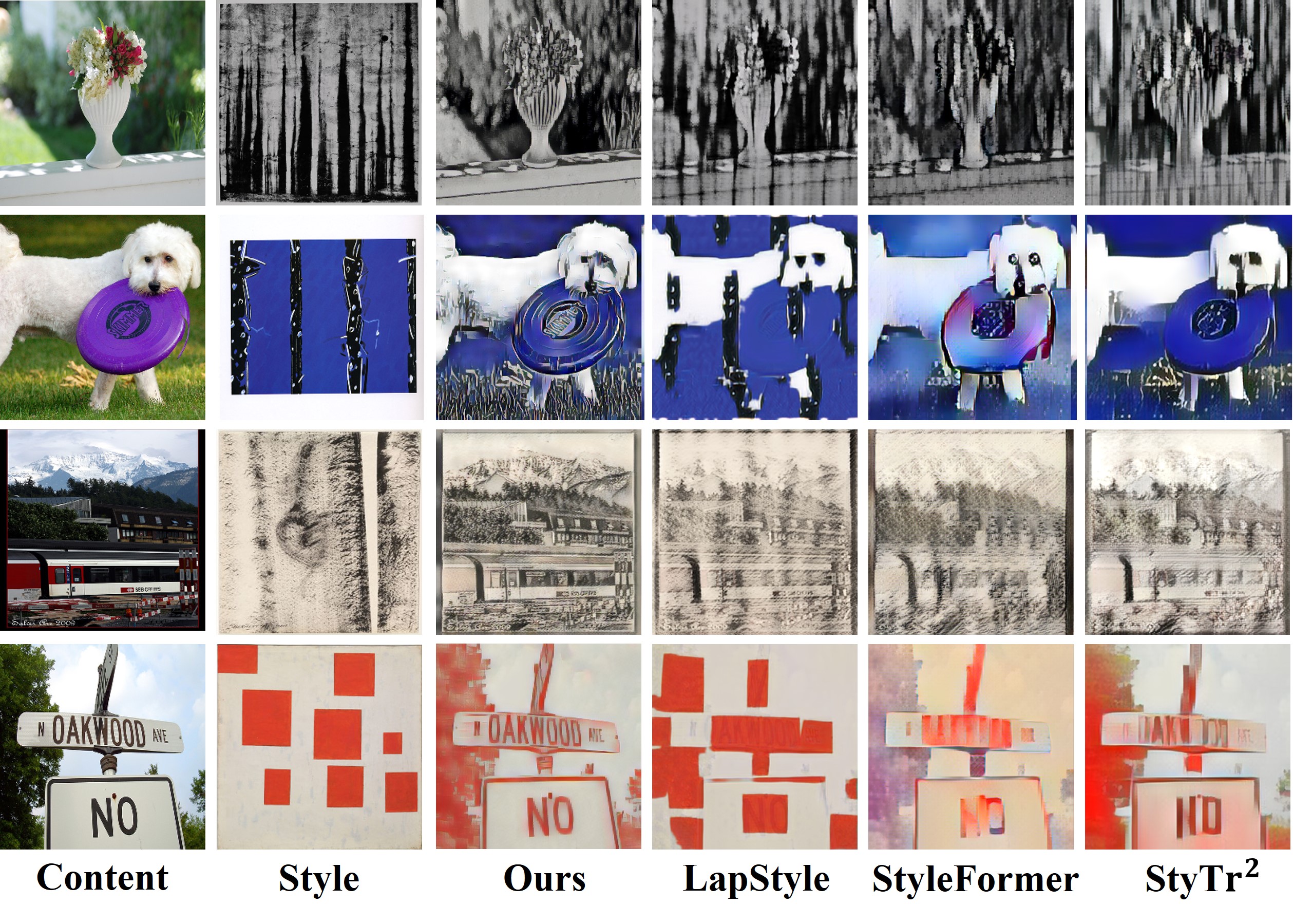}
	\end{center}
	\caption{Qualitative comparisons with LapStyle, StyleFormer and StyTr$^2$.}
	\label{fig:sota_com}
\end{figure}

\begin{figure}[t]
  \begin{center}
    \includegraphics[width=0.8\columnwidth]{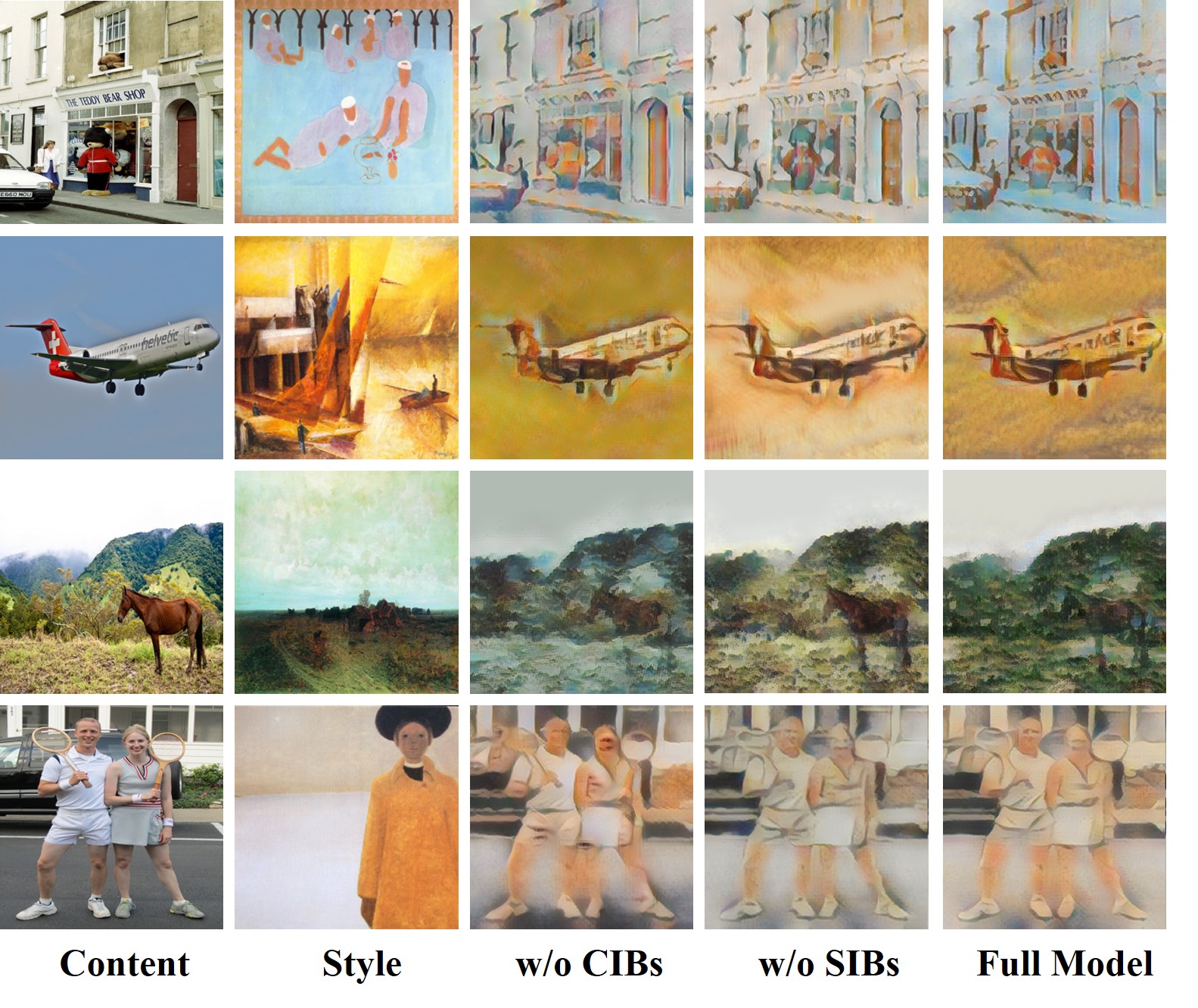}
  \end{center}
  \caption{Ablation study of the effects of the proposed CIBs and SIBs. Zoom-in for better view.}
  \label{fig:ab_ib}
\end{figure}

\subsubsection{\textbf{Quantitative Comparison}}
(a) As the evaluation of the stylization effect is quite subjective, we conduct a user study to evaluate the performance of the proposed method with other previous state-of-the-art methods. 
Firstly, we randomly sample 20 content images and 20 style images and obtain 400 generated images for each method. 
Then, we randomly give 20 generated images to 50 participants and let them select the best one in terms of three aspects: content preservation, texture stylization, and overall performance.
``Content Preservation'' evaluates the content preservation performance with respect to the photo images, while ``Texture Stylization'' evaluates the effect of style patterns with respect to the artistic images. 
``Overall Performance'' assesses the subjective perception of the quality of the generated results. 
As can be observed in Table~\ref{tab:quantitative}, our method outperforms other state-of-the-art methods in all three aspects.
Note that all the generated images are shown in random order for a fair comparison and all questionnaires are anonymous. 
(b) Furthermore, we conduct quantitative comparisons on the 400 generated images using content loss~\cite{gatys2016image}, SSIM~\cite{wang2004image}, and Gram loss~\cite{gatys2016image}. The content loss measures the similarity between the content features of the generated image and the content features of the content image. It is calculated as the mean squared error between the feature maps produced by a VGG-19 network (pretrained on ImageNet) for the content image and the generated image. SSIM stands for Structural Similarity Index Measure. It is a widely used image quality metric that measures the structural similarity between two images. It is calculated by comparing three aspects of the images: luminance, contrast, and structural similarity. It ranges from -1 to 1, with a value of 1 indicating perfect similarity between the two images. Higher SSIM values mean greater similarity between the two images. The Gram loss measures the similarity between the style features of the generated image and the style features of the style image. It is calculated as the sum of the mean squared error between the Gram matrices (a matrix of inner products of feature maps) of the VGG-19 feature maps for the style image and the generated image.  Specifically, as shown in Table~\ref{tab:quantitative}, our method achieves the best content preservation performance, reflected in the highest content loss and the lowest SSIM, thanks to the accurate disentanglement of the content structure achieved by the proposed disentangled information bottlenecks. In terms of Gram loss, WCT optimally matches the second-order statistics, thus achieving the lowest value. However, its stylized results are not as good as those produced by Infostyler, and the original content structure is also damaged. InfoStyler achieves the second-lowest value on Gram loss, demonstrating the ability to transfer rich textures and patterns.

\subsubsection{\textbf{Efficiency}}
The Execution time comparison of the proposed InfoStyler and other methods is reported in Table~\ref{tab:quantitative}. 
Two image scales are used during evaluation: 256 $\times$ 256 and 512 $\times$ 512 resolutions. 
All experiments are conducted on a single NVIDIA TITAN RTX GPU. 
Since our method use the multi-level features~(from $Relu\_2\_1$ to $Relu\_5\_1$) to produce the informative controllers and further perform style transfer on multi layers~(from $Relu\_2\_1$ to $Relu\_5\_1$), it is slightly less efficient than other methods. 
However, our model is 3-6 times faster than the matrix computation-based methods, \emph{i.e.}, WCT~\cite{li2017universal} and Avatar-Net~\cite{sheng2018avatar}.
\begin{figure}[!t]
  \begin{center}
    \includegraphics[width=0.8\columnwidth]{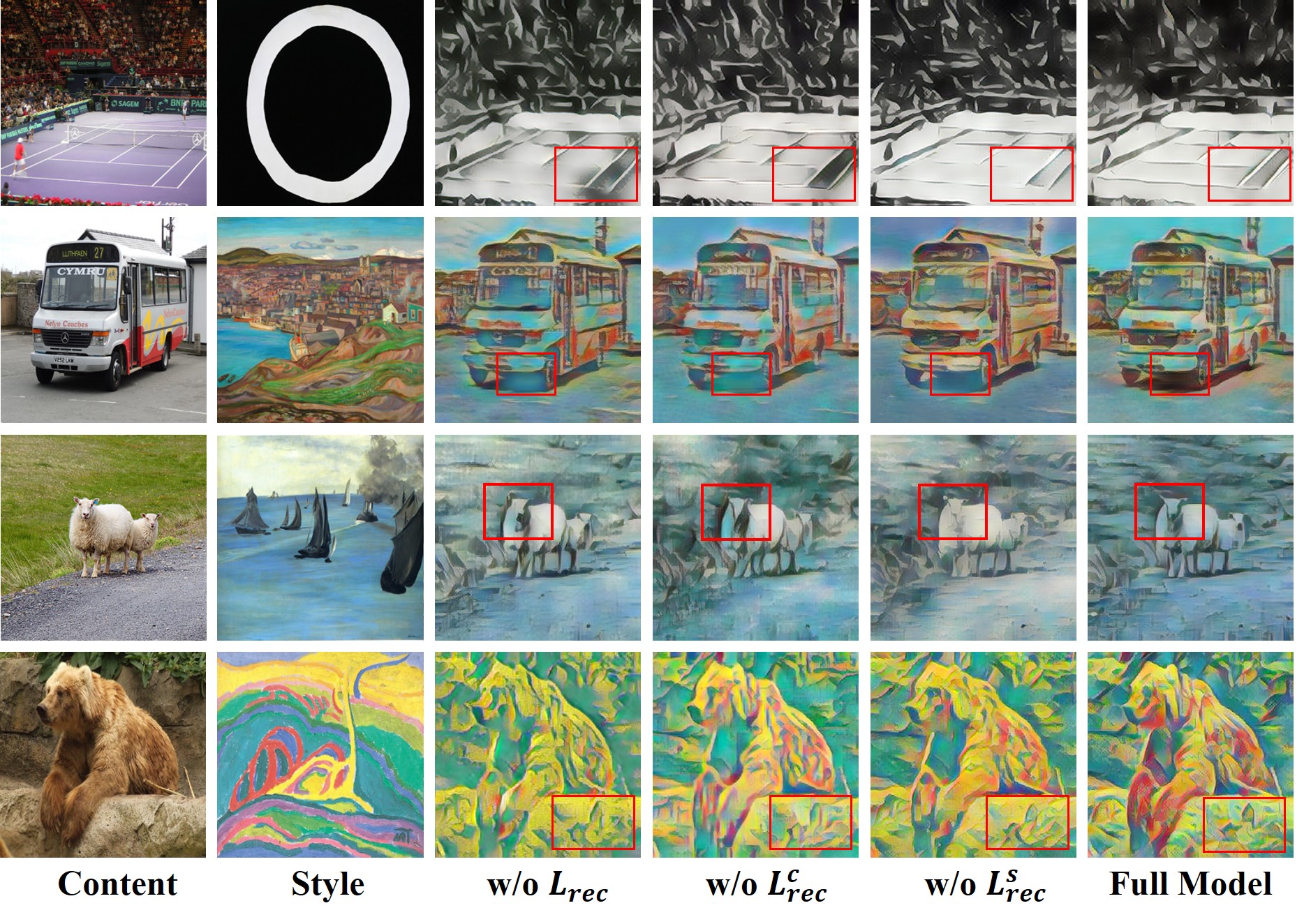}
  \end{center}
  \caption{Ablation study of the effects of the cross-domain IB learning strategy. Zoom-in for better view.}
  \label{fig:ab_cross}
\end{figure}

\begin{table}[t]
  
  \caption{Quantitative comparisons with LapStyle, StyleFormer, and StyTr$^2$.}
  \begin{center}
  \vspace{-1ex}
  \label{tab:sota_com}
    \resizebox{0.9\columnwidth}{!}{
        \begin{tabular}{lccc}
        \toprule
         & Content loss$\downarrow$ & SSIM$\uparrow$ & Gram loss$\downarrow$ \\
        \hline 
        LapStyle & $5.001$  & $0.486$ & $0.0069$ \\
        StyleFormer & $4.183$  & $0.491$ & $0.0158$ \\
        StyTr$^2$ & $4.076$ & $0.516$ & $0.0096$ \\
        Ours & $4.028$ & $0.533$ & $0.0146$ \\
        \bottomrule
        \end{tabular}}
      \end{center}
\end{table}

\begin{table}[!t]
  \centering
  \caption{Ablation study using Content loss, SSIM and Gram loss on the proposed CIBs and SIBs.}
  \vspace{-1ex}
    \resizebox{0.8\columnwidth}{!}{
        \begin{tabular}{lccc}
        \toprule
         & Content loss$\downarrow$ & SSIM$\uparrow$ & Gram loss$\downarrow$ \\
        \hline 
        w/o CIBs & $4.173$  & $0.496$ & $0.0165$ \\
        w/o SIBs & $4.035$  & $0.525$ & $0.0271$ \\
        Full Model & $4.028$ & $0.533$ & $0.0146$ \\
        \bottomrule
        \end{tabular}}
  \label{tab:ssim}
\end{table}

\begin{figure*}[!htp]
	\begin{center}
	  \includegraphics[width=0.7\linewidth]{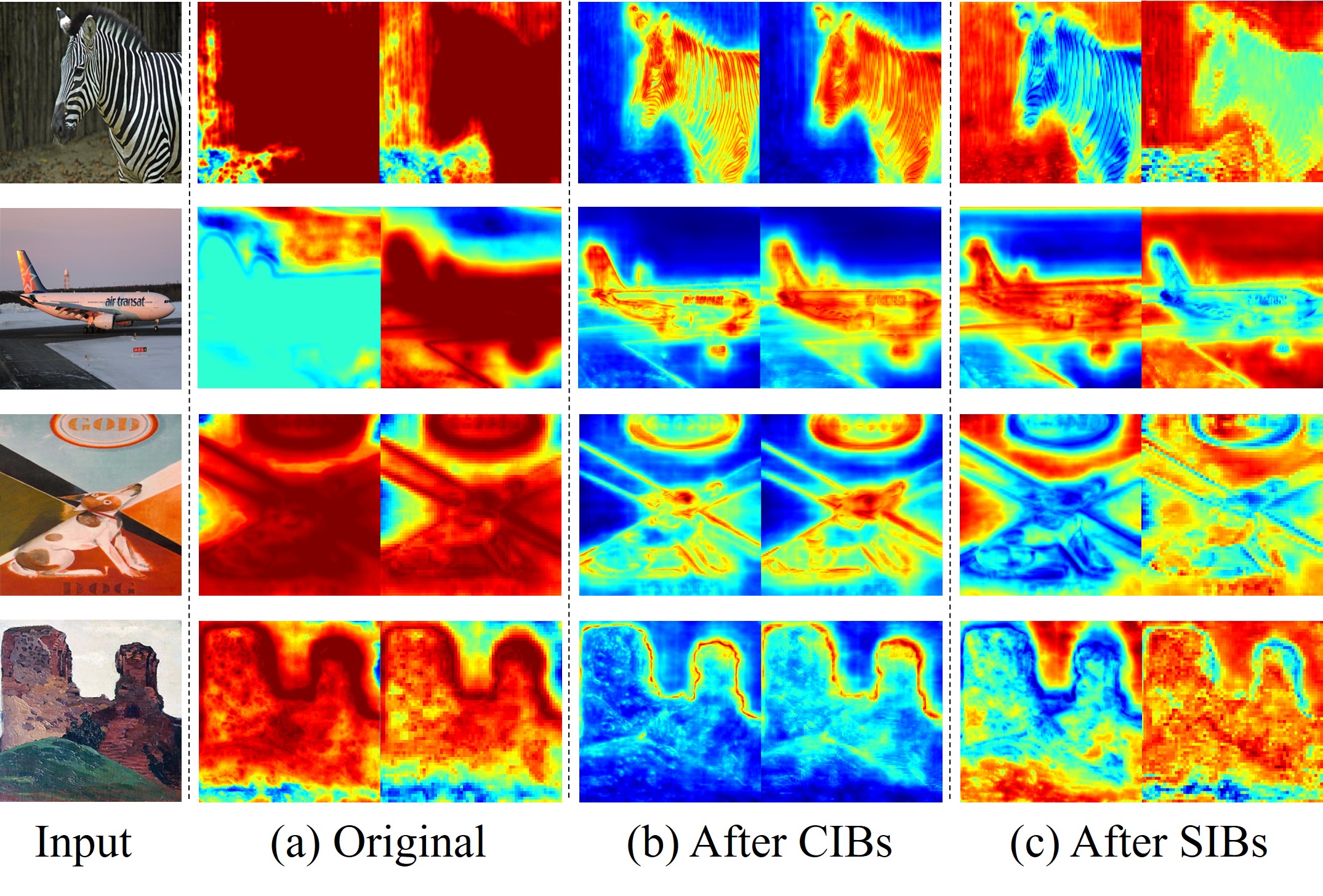}
	\end{center}
	\caption{Visualization of the information distribution of (a) the original features, and the optimized information distribution (b) after inserting CIBs and (c) after inserting SIBs in $Relu\_2\_1$ and $Relu\_3\_1$.}
	\label{fig:vis_all}
\end{figure*}

\subsection{Comparison with Current SOTA Methods}
To further verify the superiority of our proposed InfoStyler, we conduct qualitative and quantitative comparisons with current SOTA methods, LapStyle~\cite{lin2021drafting}, StyleFormer~\cite{wu2021styleformer} and StyTr$^2$~\cite{deng2022stytr2}. 
As shown in Figure~\ref{fig:sota_com}, LapStyle synthesizes some inappropriate style patterns in most cases with severe content details lost (1st, 2nd, 4th rows). This is because one LapStyle model is only optimized to a pre-defined style and easy to obtain unbalanced and over-fitting results. StyleFormer adopts the multi-head attention mechanism into the CNN-based encoder-decoder network to globally capture the style representations. However, the color distribution of stylized results is not accurate enough (2nd, 4th rows). StyTr$^2$ leverages a pure transformer-based network to perform style transfer. The style details are well transferred, but the color distributions of the content image are often mistakenly maintained (2nd, 4th rows). 
Our InfoStyler based on the disentangled information bottlenecks can preserve the content structures of the content images and achieve accurate color distribution of the given styles.
For quantitative evaluation, we randomly select 20 content images and 20 style images to obtain 400 stylized results of each method. The results are shown in Table~\ref{tab:sota_com}. Our method achieves the lowest content losses and the highest SSIM, which demonstrates the superiority of our method in content structure preservation. 
For the evaluation of texture stylization, LapStyle achieves the lowest Gram losses and our method is the second best. 
However, as discussed in the qualitative comparison above, the Gram losses of LapStyle are the lowest since it transfers local style patterns into the results excessively. Overall, our method can achieve good performance in maintaining the input content and capturing the given style.

\subsection{Model Analysis}
\subsubsection{\textbf{Effectiveness of the proposed CIBs and SIBs}}
The proposed CIBs and SIBs aim to disentangle content-related features and style-related features explicitly. 
They work together to prompt the network to generate more robust and accurate stylized results.
We analyze the effect of the proposed CIBs and SIBs in this part. 

As shown in Figure~\ref{fig:ab_ib}, we first conduct an ablation study by removing the CIBs or SIBs, respectively.
Without the CIBs, the content structures in both results are less preserved. For example, the structure of the doll in row 1 and the shape of the plane in row 2 become blurred and unclear.
Without the SIBs, though the content structures can be preserved to some extent, we can observe less stylized results are produced. 
By combining CIBs and SIBs together, the InfoStyler can flexibly represent the style information, while preserving the content structure. 
As shown in Figure~\ref{fig:vis_all}, we then visualize the information distributions in two intermediate layers, \emph{i.e.}, $Relu\_2\_1$ and $Relu\_3\_1$.
Specifically, the columns (a) show the original uncompressed information distribution in $Relu\_2\_1$ and $Relu\_3\_1$. The columns (b) and (c) illustrate the compressed information distribution after inserting CIBs and SIBs in $Relu\_2\_1$ and $Relu\_3\_1$, respectively.
We can observe that the original uncompressed VGG-based information distribution is almost constant in all regions indiscriminately. After introducing CIBs, the content information is accurately compressed to areas that reflect structural information such as edges and contours. 
While after introducing SIBs, the style information is optimized to regions with rich styles such as color patterns, and texture patterns.

We further show quantitative results in Table~\ref{tab:ssim}. Specifically, we randomly select 20 content images and 20 style images to obtain 400 stylized results for different settings. We use the Structural Similarity Index Measure (SSIM) and the content loss between the content images and the stylized images to measure the performance of content structure preservation. Inspired by \cite{li2017universal}, we adopt the Gram matrices between the style images and the transferred images to evaluate the stylization effects. We can observe that without CIBs, the content preservation performance is degraded, as reflected by higher content loss and lower SSIM metric. Without SIBs, the stylized effect is weakened and the gram loss is higher. By combining CIBs and SIBs, our method achieves the best performance in all evaluations, which verifies that it can well preserve the content structure while flexibly representing rich styles. The quantitative comparison leads to the same conclusion as the visual comparison in Figure~\ref{fig:ab_ib}.
\begin{figure}[t]
  \begin{center}
  \includegraphics[width=1\columnwidth]{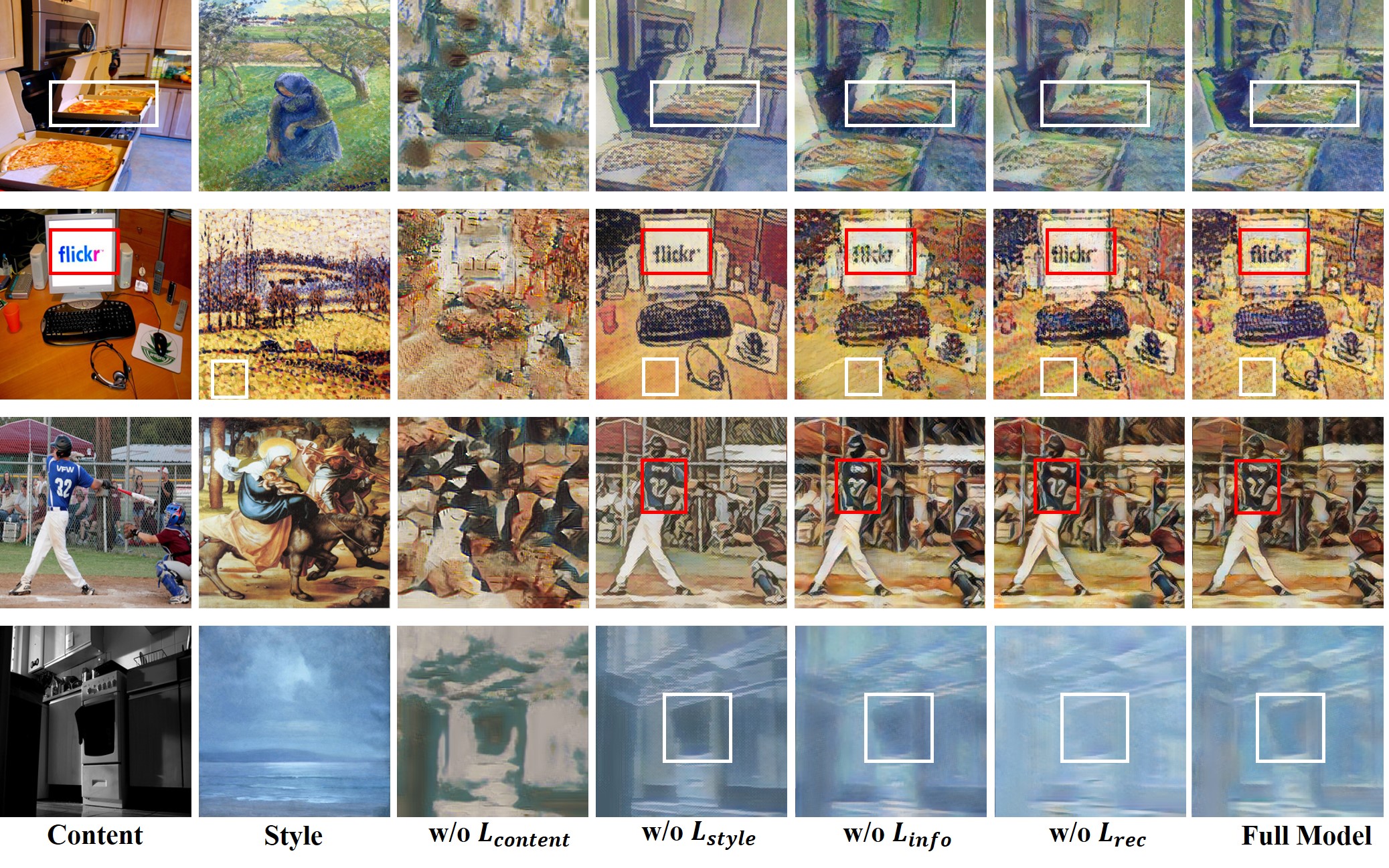}
  \end{center}
  \caption{Ablation study on each loss term. Zoom in for better view.}
  \label{fig:abs_loss}
  \end{figure}
    
\begin{table}[t]

\caption{Ablation study on different loss terms evaluated by Content loss, SSIM and Gram loss.}
\begin{center}
\label{tab:ab_loss}
\resizebox{0.8\columnwidth}{!}{
  \begin{tabular}{lccc}
  \toprule & Content loss$\downarrow$ & SSIM$\uparrow$ & Gram loss$\downarrow$ \\
  \hline 
  w/o $L_{content}$ & $6.927$  & $0.408$ & $0.0282$ \\
  w/o $L_{style}$ & $3.955$  & $0.548$ & $0.0448$ \\
  w/o $L_{info}$ & $4.212$  & $0.503$ & $0.0192$ \\
  w/o $L_{rec}$ & $4.10$  & $0.511$ & $0.0166$ \\
  Full Model & $4.028$ & $0.533$ & $0.0146$ \\
  \bottomrule
  \end{tabular}}
\end{center}
\end{table}

\begin{figure}[!t]
  \begin{center}
  \includegraphics[width=0.75\columnwidth]{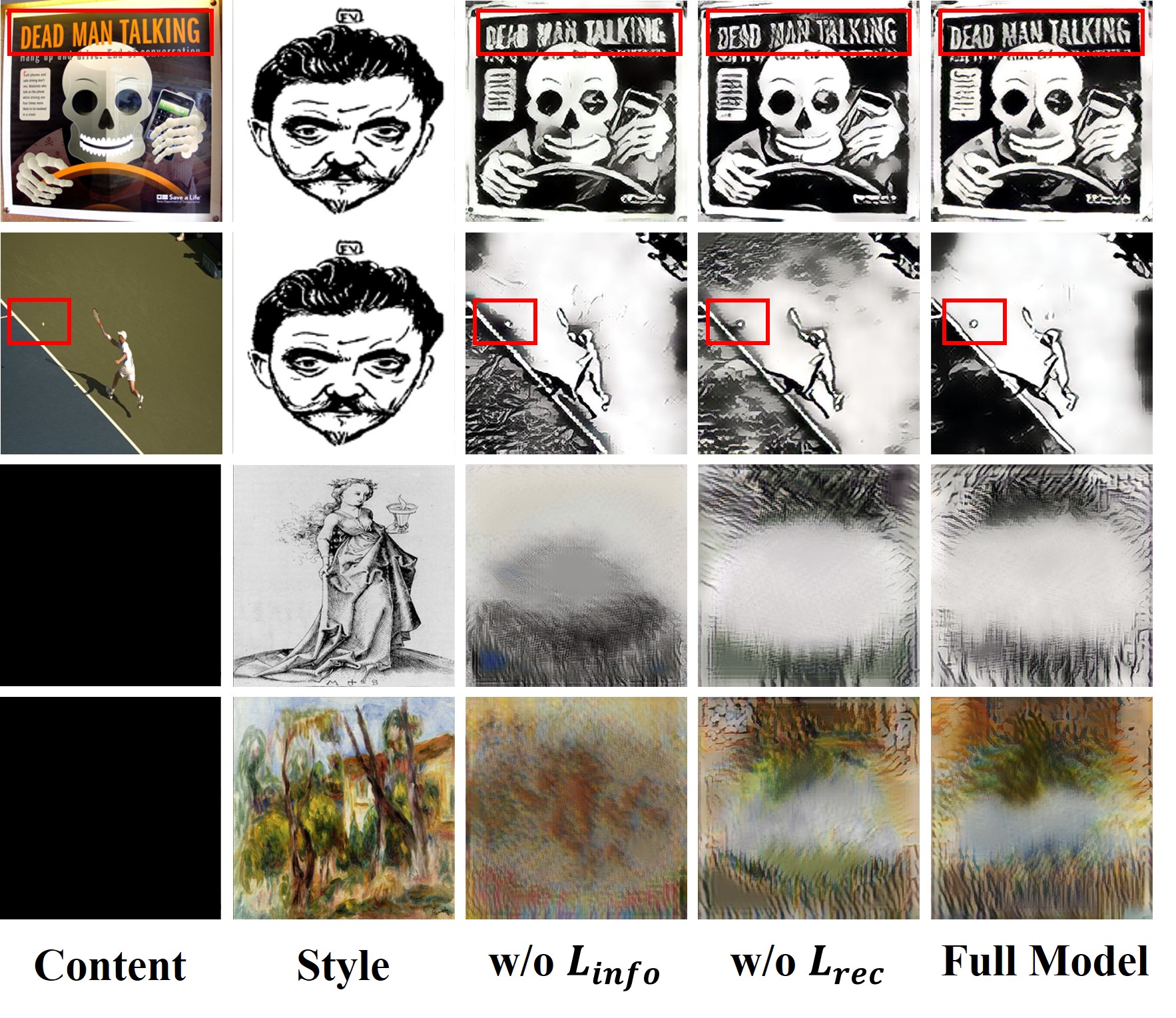}
  \end{center}
  \caption{Evaluation of disentanglement performance by using a simple style image, such as a black-line drawing, as style input to evaluate the disentanglement of content information (shown in the first two rows), and using a simple content image, such as a black image, as content input to evaluate the disentanglement of style information (shown in the last two rows).}
  \label{fig:dis}
\end{figure}
  
  \begin{table}[t]
    
    \caption{The values of the mutual information before and after injecting noises in the internal content and style features with or without the supervision of $\mathcal{L}_{info}$.}
    \begin{center}
    \label{tab:mi}
    \resizebox{0.9\columnwidth}{!}{
    \begin{tabular}{ll|l|l|l|l}
    \toprule
    \multicolumn{2}{l|}{} & $Relu\_2\_1$ & $Relu\_3\_1$ & $Relu\_4\_1$ & $Relu\_5\_1$ \\ 
    \hline
    \hline
    \multicolumn{1}{l|}{\multirow{2}{*}{CIBs}} & w/o $\mathcal{L}_{info}$ & 2.773& 1.958 & 2.916& 2.139\\ \cline{2-6} 
    \multicolumn{1}{l|}{}                      & w $\mathcal{L}_{info}$   & 0.329& 0.169 & 0.025& 0.001\\ \hline
    \multicolumn{1}{l|}{\multirow{2}{*}{SIBs}} & w/o $\mathcal{L}_{info}$ & 1.250& 2.454 & 2.532& 1.727\\ \cline{2-6} 
    \multicolumn{1}{l|}{}                      & w $\mathcal{L}_{info}$   & 0.981& 1.338 & 1.136& 0.995\\ 
    \bottomrule
    \end{tabular}}
    \end{center}
  \end{table}

\subsubsection{\textbf{Effectiveness of Cross-domain IB Learning}}
As demonstrated in Figure~\ref{fig:ab_cross}, we conduct ablation study under different configurations: 
(1) ``w/o $L_{rec}$'' represents the cross-domain IB learning strategy is not performed during training.
(2) ``w/o $L_{rec}^{c}$'' represents performing the cross-domain IB learning without content reconstruction.
(3) ``w/o $L_{rec}^{s}$'' indicates performing the cross-domain IB learning without style reconstruction.
(4) ``Full Model'' indicates performing the cross-domain IB learning with both domains reconstruction.
When ``w/o $L_{rec}$'', the transferred results are impressive but sometimes there are unnatural transferred areas in the results, such as the styles near the edges that make the edges unclear, which is labeled in Figure~\ref{fig:ab_cross}.
When ``w/o $L_{rec}^{c}$'', the feedforward network only need to reconstruct the style domain $\mathcal{S}$.
Since the style statistics to be reconstructed are extracted from the stylized output $I_{cs}$, it prompts containing more style information in the stylization results but may ignore the importance of content preservation, which can be shown in the fourth column of Figure~\ref{fig:ab_cross}. 
On the contrary, under the configuration of ``w/o $L_{rec}^{s}$'', the network is only supervised by the reconstruction of the content domain $\mathcal{C}$. 
Therefore, the network is biased to preserve the content structure and may produce less stylized results with limited texture patterns, which is shown in the fifth column of Figure~\ref{fig:ab_cross}. 
As shown in the last column of Figure~\ref{fig:ab_cross}, especially in the marked regions, the full model can better balance content structure preservation and style pattern transfer, obtaining impressive results,
which demonstrates the effectiveness of cross-domain IB learning by reconstructing the content and style domains. 
\begin{figure*}[htb]
  \begin{center}
    \includegraphics[width=0.8\linewidth]{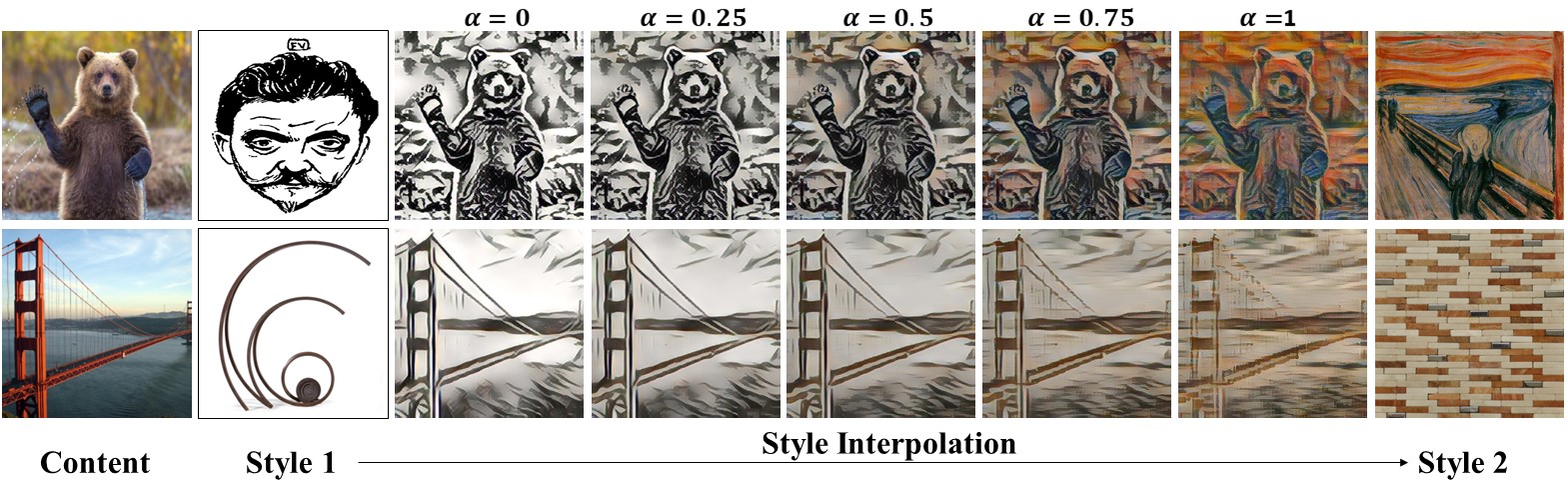}
  \end{center}
  \caption{Results of style interpolation. The styles of the interpolated images continuously transfer from style 1 to style 2 while simultaneously preserving the content information.}
  \label{fig:inter}
\end{figure*}

\begin{table}[t]
  
  \caption{Quantitative evaluation by using the content loss and SSIM for content information disentanglement and using the Gram loss for style information disentanglement.}
  \begin{center}
  \label{tab:dis}
  \resizebox{0.9\columnwidth}{!}{
  \begin{tabular}{cc|c|c|c}
  \toprule
  \multicolumn{2}{l|}{}  & Content loss $\downarrow$ & \multicolumn{1}{l|}{SSIM $\uparrow$} & Gram loss $\downarrow$ \\ 
  \hline
  \hline
  \multicolumn{1}{c|}{\multirow{3}{*}{\begin{tabular}[c]{@{}c@{}}Disentanglement\\ for content information\end{tabular}}}& w/o $\mathcal{L}_{info}$ & 4.223 & 0.445 & -\\ \cline{2-5} 
  \multicolumn{1}{c|}{}& w/o $\mathcal{L}_{rec}$ & 4.154& 0.451 & - \\ \cline{2-5} 
  \multicolumn{1}{c|}{}& Full Model & 3.991 & 0.460 & - \\ \hline
  \multicolumn{1}{c|}{\multirow{3}{*}{\begin{tabular}[c]{@{}c@{}}Disentanglement\\ for style information\end{tabular}}}& w/o $\mathcal{L}_{info}$   & -  & -   & 0.0219     \\ \cline{2-5} 
  \multicolumn{1}{c|}{}& w/o $\mathcal{L}_{rec}$ & -   & - & 0.0147     \\ \cline{2-5} 
  \multicolumn{1}{c|}{}                                                                                                   & \multicolumn{1}{l|}{Full Model} & -  & -   & 0.0121     \\ \bottomrule
  \end{tabular}}
  \end{center}
  \end{table}

\subsubsection{\textbf{{Effectiveness of different loss terms}}}
We conduct ablation study to verify the effectiveness of different loss terms used for InfoStyler.
We show the results by removing each loss term in Figure~\ref{fig:abs_loss}. 
(1) Without $\mathcal{L}_{content}$, the visual quality is significantly worse than that of the full model. Even though the color distribution is transferred to some extent, the original content structures are totally damaged. 
(2) Without $\mathcal{L}_{style}$, the style patterns (\emph{e.g.}, colors) are still weakly transferred, since the adopted content loss uses the initial transfer results as the content supervision in Eq.~\ref{con:content}. However, the overall color saturation is degraded, and the color distribution of local regions (shown in white boxes) is wrongly maintained from the content images. 
(3) Without $\mathcal{L}_{info}$, 
the content structures are not well preserved (shown in the red boxes) and the transferred style patterns are not accurate for the given styles (shown in the white boxed), as the original VGG-based features are not optimized.
(4) Without $\mathcal{L}_{rec}$, the visual quality is still worse than that of the full model. 
The combination of these four losses (Full Model) results in generated images that have a balance between content structure preservation and style pattern richness. 

We show quantitative results in Table~\ref{tab:ab_loss}. Specifically, we randomly select 20 content images and 20 style images to obtain 400 stylized results for evaluation. It can be observed that the results without $\mathcal{L}_{content}$ have the highest Content loss and the lowest SSIM, illustrating the poorest performance on content structure preservation. The results without $\mathcal{L}_{style}$ have the highest Gram loss, showing the weakest stylization performance. 
The results without $\mathcal{L}_{info}$ still have poorer performance than the results of the full model. By introducing $\mathcal{L}_{info}$ and $\mathcal{L}_{rec}$, the results of the full model have the lowest Content loss, the highest SSIM index, and the lowest Gram loss, obtaining the best style transfer performance.
  
\subsubsection{\textbf{{Information compression analysis}}}
To verify the performance on information compression of InfoStyler, we calculate the average values of mutual information in the CIB and SIB modules at different layers. As shown in Table~\ref{tab:mi}, without the supervision of $\mathcal{L}_{info}$, the values of mutual information at each layer are high. After introducing $\mathcal{L}_{info}$ to supervise the reduction of the mutual information before and after injecting noises, the values of mutual information at each layer are reduced. Note that for CIB modules, the reduction in mutual information values is greater than that of the SIB modules. This is because the CIB modules require introducing more noises to make the content encoding network only focus on the structural information in small regions, which is consistent with the findings in Figure~\ref{fig:vis_all}.
\begin{figure}[t]
  \begin{center}
    \includegraphics[width=0.8\columnwidth]{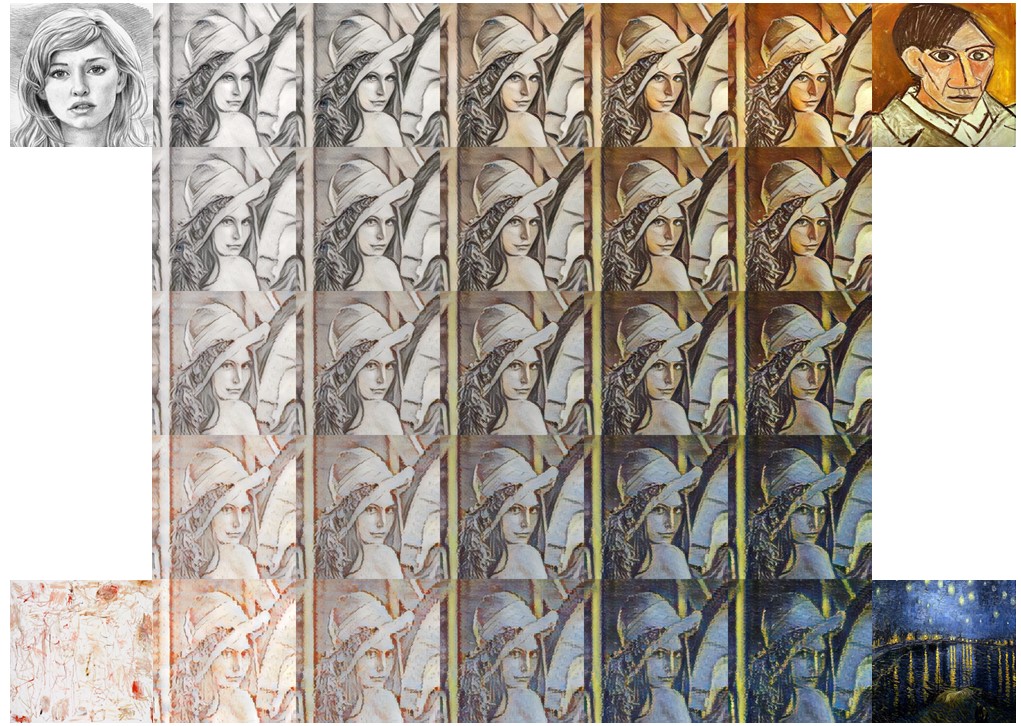}
  \end{center}
  \caption{Results of style interpolation with four different styles.}
  \label{fig:multiinter}
\end{figure}

\subsubsection{\textbf{{Disentanglement analysis}}}
To evaluate the disentanglement performance of our method, we propose an intuitive disentanglement evaluation method and compare the performance of models without $\mathcal{L}_{info}$, without $\mathcal{L}_{rec}$, and the full model. Through the disentanglement evaluation, we can verify the importance of the proposed IB learning and cross-domain training strategy for disentangled style transfer. Specifically, to evaluate the disentanglement for content information, we select an image with only black-lines as the style image, and evaluate whether the generated line-art results can ensure the structure of the original content. To evaluate the disentanglement for style information, we select a black image as the content image and evaluate whether the transferred style on the black image can follow the original style distribution.

The results are shown in Figure~\ref{fig:dis}. The first two rows show the results with a black-line image as the style image. Without $\mathcal{L}_{info}$, the structure of the generated results is somewhat damaged (as shown in the red boxes), which means poor disentanglement for content information without information compression. 
Without $\mathcal{L}_{rec}$, the content structure of the generated results remains better, indicating cross-domain IB learning can further improve the content preservation. The full model best maintains the structure by introducing both losses. The last two rows show the results with a black image as the content image. It can be seen that without $\mathcal{L}_{info}$, the color distribution of the generated results is largely different from the original style, indicating the necessity of $\mathcal{L}_{info}$ for accurate style disentanglement. By introducing $\mathcal{L}_{info}$, the generated style and the original style distribution are pulled closer, for both without $\mathcal{L}_{rec}$ and the full model.

Furthermore, we conduct quantitative evaluations on the generated results. Specifically, we randomly selected 400 images with different contents and the black-line style image for content preservation-related evaluation. We also randomly select 400 images with different styles and a black content image for style preservation-related evaluation. As shown in Table~\ref{tab:dis}, all matrices achieve significant improvements after introducing $\mathcal{L}_{info}$ and $\mathcal{L}_{rec}$, with $\mathcal{L}_{info}$ showing a more obvious enhancement in disentanglement performance. This indicates that our method can achieve good disentanglement through information compression, and improve the performance of style transfer.

\subsubsection{\textbf{{Interpolated Style Transfer}}}
To show the smoothness of the learned model, we can also achieve style interpolation between a set of $N$ reference images, $I^1_s,I^2_s,\cdots,I^N_s$, with corresponding weights $\alpha_1,\alpha_2,\cdots,\alpha_N$ such that $\sum^N_{n=1}\alpha_n=1$. As shown in Figure~\ref{fig:inter}, we achieve natural and smooth interpolated results from $I^1_s$ to $I^2_s$. As illustrated in Figure~\ref{fig:multiinter}, the results of style interpolation with four different style images, $I^1_s,I^2_s,I^3_s,I^4_s$, are shown. Although four feature maps from different reference images are combined, the image quality of the results is always high and the results are natural.

\section{Conclusions}
In this paper, we have proposed a novel information disentanglement approach, named \textit{InfoStyler}, to capture disentangled content and style representation from the perspective of information theory. 
InfoStyler formulates the disentanglement representation learning as an information compression problem by extracting the minimal sufficient information for both content and style representations from the pre-trained encoding network. 
Besides, to further facilitate the representation disentanglement, a cross-domain Information Bottleneck (IB) learning strategy is introduced to reconstruct the content domain and the style domain in a symmetric cycle manner. 
The qualitative and quantitative experiments show that InfoStyler achieves superior stylized results while balancing content structure preservation and style pattern richness.
Also, through visualization experiments, the well-disentangled representation of the proposed InfoStyler is verified. 
In future work, we will explore its usage and possible variants in other image manipulation or translation tasks. 

\section{Acknowledgements} 
This work is supported by the National Key Research and Development Program of China under Grant No. 2021YFC3320103, the National Natural Science Foundation of China (NSFC) under Grants 62272460, U19B2038, Beijing Natural Science Foundation under Grant No. 4232037, and a grant from Young Elite Scientists Sponsorship Program by CAST (YESS).

\bibliographystyle{IEEEtran}
\bibliography{egbib}

\begin{IEEEbiography}[{\includegraphics[width=1in,height=1.25in,clip,keepaspectratio]{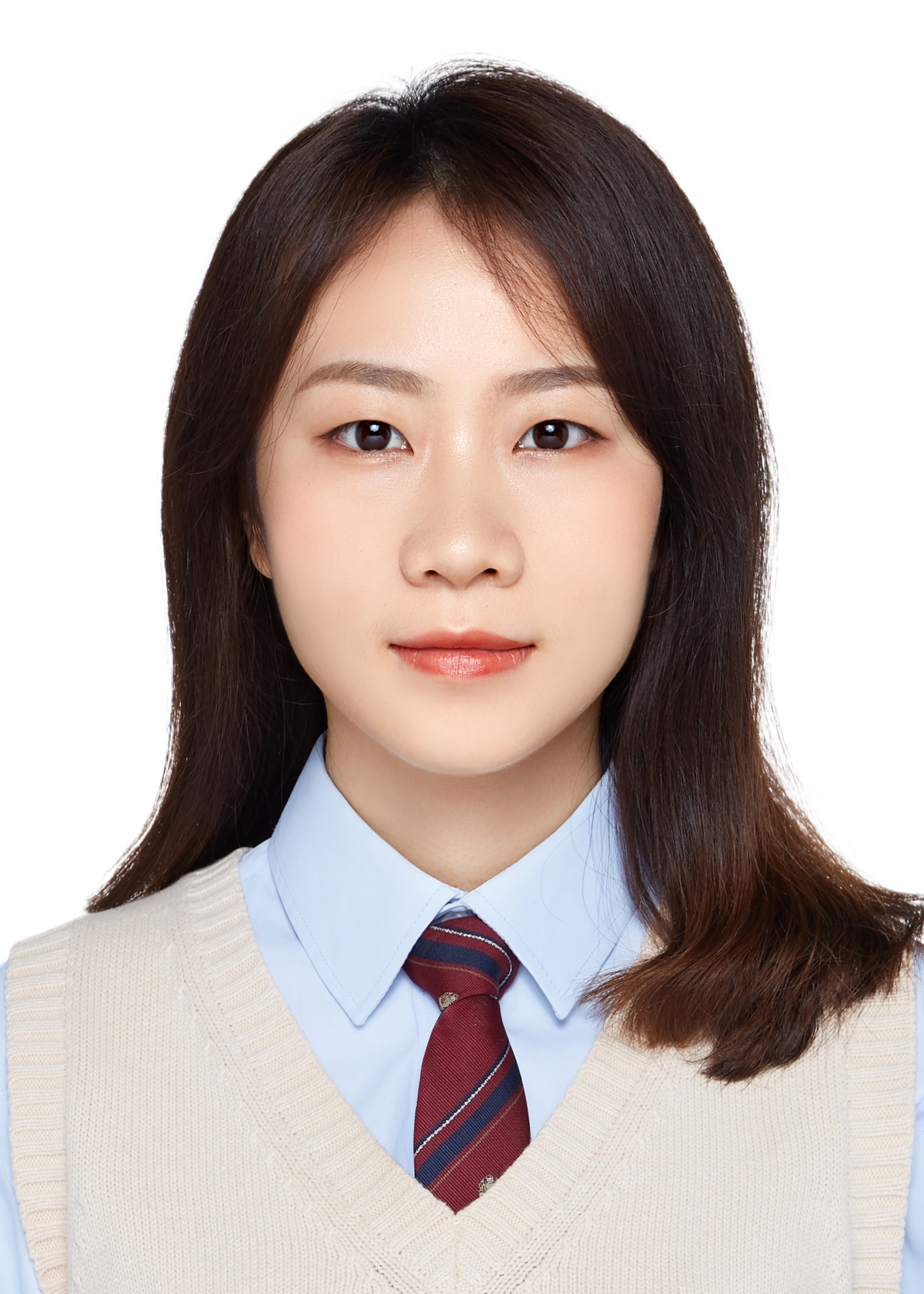}}]{Yueming Lyu}{\space}received B.Eng. degree in Nanjing University of Aeronautics and Astronautics, China in 2019. She is a Ph.D. degree candidate in the Center for Research on Intelligent Perception and Computing (CRIPAC) at the State Key Laboratory of Multimodal Artificial Intelligence Systems, Institute of Automation, Chinese Academy of Sciences, China. Her current research include computer vision, image generation and adversarial learning.

\end{IEEEbiography}

\begin{IEEEbiography}[{\includegraphics[width=1in,height=1.25in,clip,keepaspectratio]{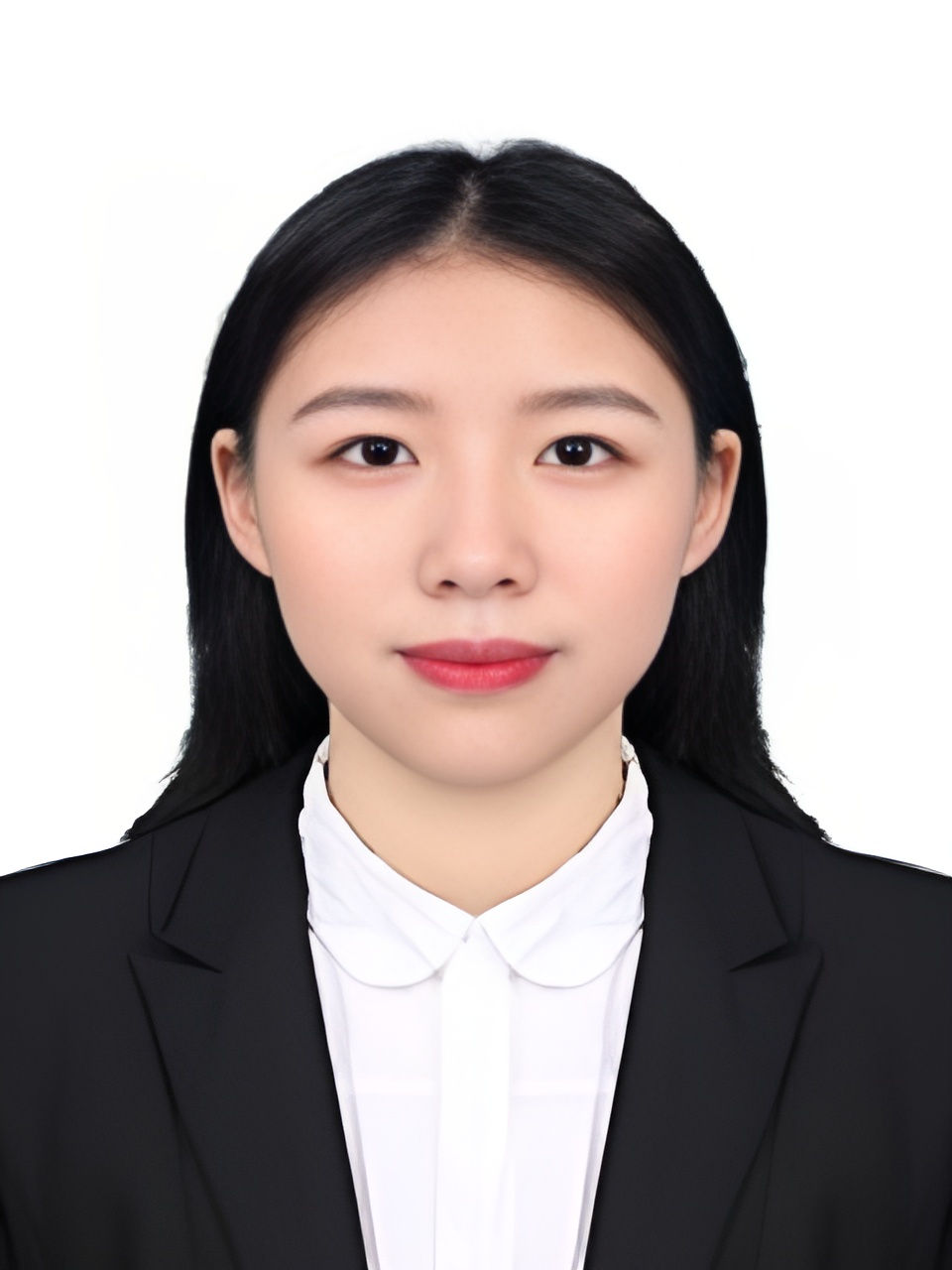}}]{Yue Jiang}{\space}received B.Eng. degree in Xi'an Jiaotong University, China in 2021. She is a Master degree candidate in the Center for Research on Intelligent Perception and Computing (CRIPAC) at the State Key Laboratory of Multimodal Artificial Intelligence Systems, Institute of Automation, Chinese Academy of Sciences, China. Her current research focuses on computer vision and image generation.

\end{IEEEbiography}

\begin{IEEEbiography}[{\includegraphics[width=1in,height=1.25in,clip,keepaspectratio]{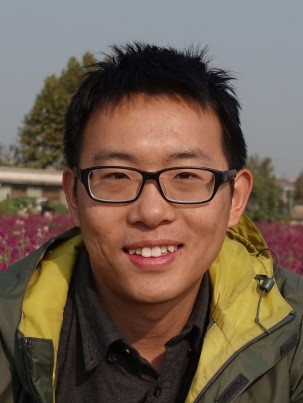}}]{Bo Peng}{\space}received B.Eng. degree in Beihang University and PhD degree in the Institute of Automation, Chinese Academy of Sciences in 2013 and 2018, respectively. Since July 2018, Dr. Bo Peng has joined the Institute of Automation, Chinese Academy of Sciences where he is currently an Associate Professor. His current research focuses on computer vision and image forensics. 

\end{IEEEbiography}

\begin{IEEEbiography}[{\includegraphics[width=1in,height=1.25in,clip,keepaspectratio]{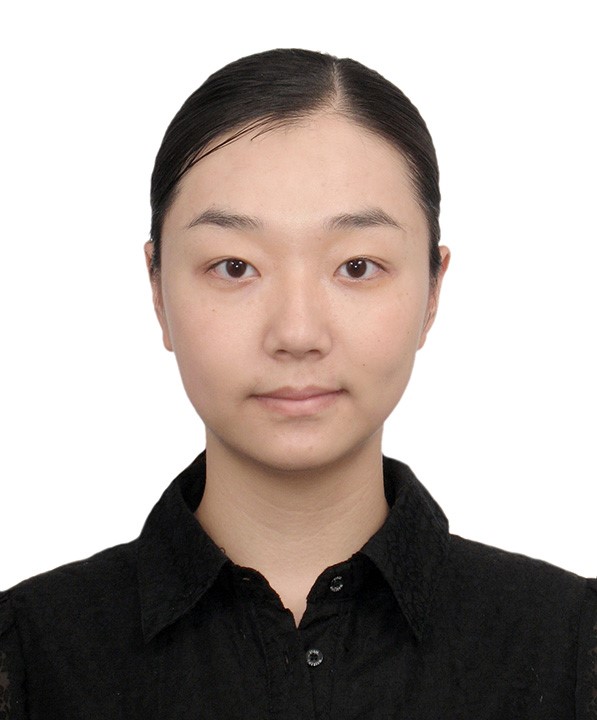}}]{Jing Dong}{\space}recieved her Ph.D in Pattern Recognition from the Institute of Automation, Chinese Academy of Sciences, China in 2010. Then she joined the Institute of Automation, Chinese Academy of Sciences and she is currently a Professor. Her research interests are towards Pattern Recognition, Image Processing and Digital Image Forensics including digital watermarking, steganalysis and tampering detection. She is a senior member of IEEE. She also has served as the deputy general of Chinese Association for Artificial Intelligence.

\end{IEEEbiography}
\vfill

\end{document}